\documentclass{article}

\usepackage[final]{neurips_2025}

\usepackage[utf8]{inputenc} 
\usepackage[T1]{fontenc}    
\usepackage{hyperref}       
\usepackage{url}            
\usepackage{booktabs}       
\usepackage{amsfonts}       
\usepackage{nicefrac}       
\usepackage{microtype}      
\usepackage{xcolor}         
\usepackage{amsmath}
\usepackage{graphicx}
\usepackage{subfigure}
\usepackage{enumitem}
\usepackage{algorithm}
\usepackage{algorithmic}
\usepackage{subcaption}
\usepackage{caption}
\usepackage{wrapfig}
\usepackage{comment}

\usepackage{amssymb}
\usepackage{mathtools}
\usepackage{amsthm}
\usepackage{pifont}
\usepackage{upgreek}
\usepackage{array}
\usepackage{ulem}
\usepackage[capitalize,noabbrev]{cleveref}
\usepackage{multirow}

\hypersetup{
    colorlinks=true,          
    citecolor=blue,           
    allcolors=blue,           
    pdfborderstyle={/S/U/W 1} 
}

\title{{\it HiMaCon:} Discovering Hierarchical Manipulation Concepts from Unlabeled Multi-Modal Data}

\author{
Ruizhe Liu$^{1}$\ \ ~ 
Pei Zhou$^{1}$\ \ ~ 
Qian Luo$^{1,4}$\ \ ~
Li Sun$^{1}$\ \ ~
\\
\textbf{Jun Cen}$^{3}$\ \ ~ 
\textbf{Yibing Song}$^{3}$\ \ ~ 
\textbf{Yanchao Yang}$^{1,2}$
\\
\\
{$^{1}$HKU Musketeers Foundation Institute of Data Science, The University of Hong Kong} \\
{$^{2}$Department of Electrical and Electronic Engineering, The University of Hong Kong} \\
{$^{3}$DAMO Academy, Alibaba Group}\quad
{$^{4}$Transcengram}\\
{\texttt{\{zrllrz360,pezhou,qianluo,sunlids\}@connect.hku.hk}}\\
{\texttt{\{cenjun.cen,songyibing.syb\}@alibaba-inc.com},
\texttt{yanchaoy@hku.hk}}
}

\begin{document}

\maketitle

\vspace{-5mm}
\begin{abstract}
Effective generalization in robotic manipulation requires representations that capture invariant patterns of interaction across environments and tasks.
We present a self-supervised framework for learning hierarchical manipulation concepts that encode these invariant patterns through cross-modal sensory correlations and multi-level temporal abstractions without requiring human annotation.
Our approach combines a cross-modal correlation network that identifies persistent patterns across sensory modalities with a multi-horizon predictor that organizes representations hierarchically across temporal scales. Manipulation concepts learned through this dual structure enable policies to focus on transferable relational patterns while maintaining awareness of both immediate actions and longer-term goals.
Empirical evaluation across simulated benchmarks and real-world deployments demonstrates significant performance improvements with our concept-enhanced policies. 
Analysis reveals that the learned concepts resemble human-interpretable manipulation primitives despite receiving no semantic supervision. This work advances both the understanding of representation learning for manipulation and provides a practical approach to enhancing robotic performance in complex scenarios.
Code is available at: \href{https://github.com/zrllrz/HiMaCon}{\texttt{https://github.com/zrllrz/HiMaCon}}.
\end{abstract}

\section{Introduction}
\label{sec:introduction}

\begin{figure}
\centering
\includegraphics[width=\linewidth]{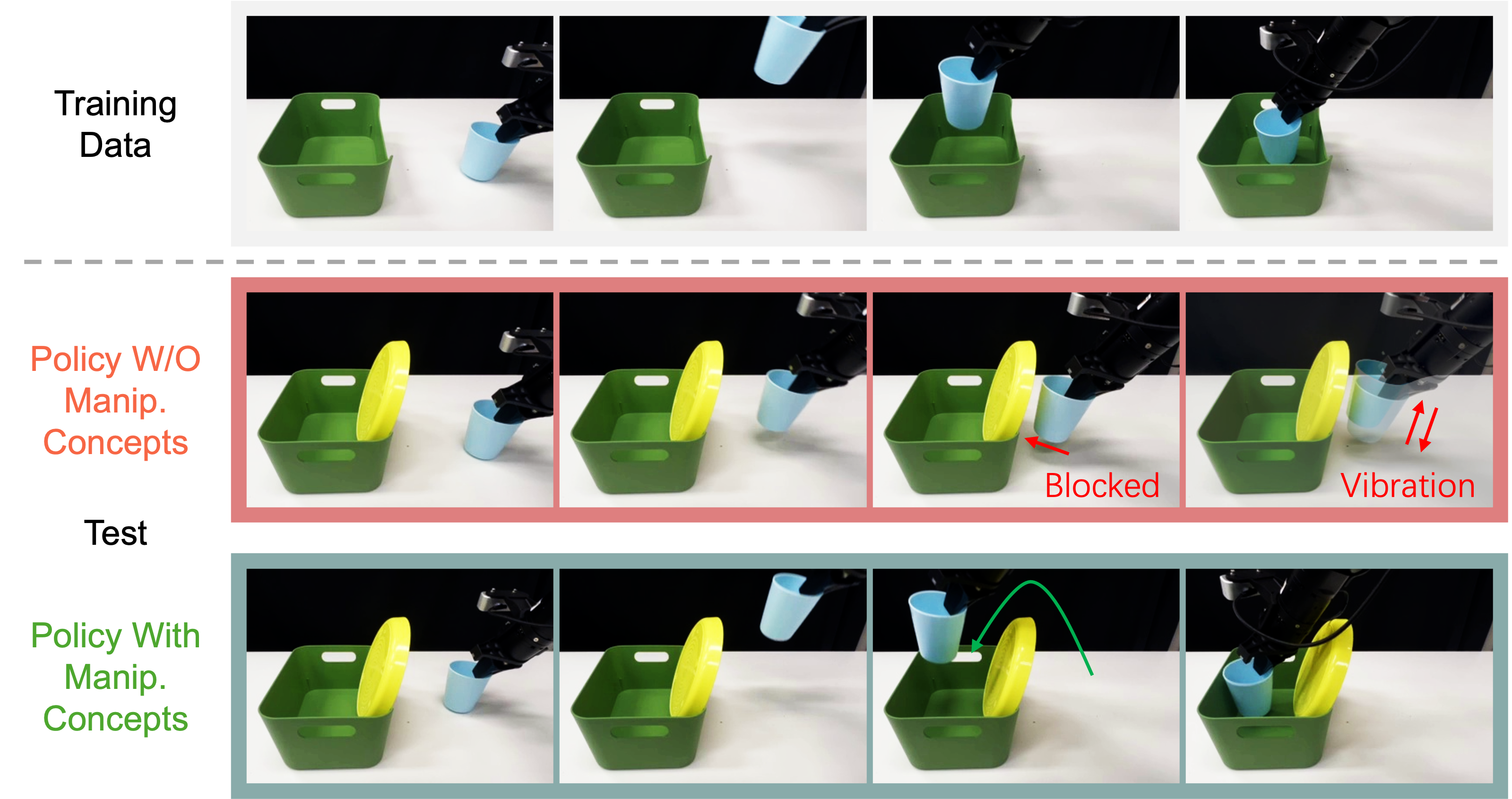}
\caption{
\textbf{Manipulation concepts enhance generalization.} 
Top: Training data with cups and containers without barriers. 
Middle: Without manipulation concepts, policies fail when encountering barriers. 
Bottom: With our concept enhancement, policies adapt accordingly.}
\label{fig:intro}
\vspace{-4mm}
\end{figure}

Robot manipulation in diverse, unstructured environments remains a fundamental challenge. 
Despite advances in 
policy learning and architectures~\citep{black2024pi_0,oxe,intelligence2025pi,kim2024openvla}, 
current approaches often fail when encountering unexpected variations or novel scenarios. 
As illustrated in Fig.~\ref{fig:intro}, 
a policy trained to place cups into containers may succeed in familiar settings but fail when encountering unexpected barriers---revealing a critical generalization gap limiting real-world deployment.

We propose that addressing this challenge requires learning transferable \textit{manipulation concepts}---hierarchical abstractions capturing fundamental manipulation patterns. 
These concepts connect low-level actions to high-level goals,
enabling robust generalization.
For example, the concept of ``placing an object inside a container'' encompasses invariant relational patterns that persist whether the container has barriers or not,
allowing adaptation while maintaining core manipulation strategy.

To acquire these manipulation concepts, 
we propose a self-supervised framework that learns hierarchical latent representations without requiring labor-intensive human annotations~\citep{eisner2022flowbot3d,li2023manipllm,mo2019partnet}. Our approach operates through two complementary mechanisms:
1) \textit{Cross-modal correlation learning} captures invariant patterns across different sensory modalities (vision, proprioception), enabling generalization across visual variations while preserving functional relationships. 
When placing objects in containers, these correlations encode the relationship between visual perception of container boundaries and proprioceptive feedback during placement, regardless of container appearance.
2) \textit{Multi-horizon sub-goal organization} structures concepts hierarchically across temporal scales, from immediate actions (e.g., ``align gripper with object'') to extended sequences (e.g., ``transport object to container''). 
This hierarchical representation enables policies to simultaneously reason about immediate actions and longer-term goals, maintaining task coherence even when specific execution paths require adaptation.

Our experiments across both simulated benchmark tasks and real-world robot deployments demonstrate that policies enhanced with these manipulation concepts consistently outperform conventional approaches, particularly in challenging scenarios requiring adaptation to novel objects, unexpected obstacles, and environmental variations (Fig.~\ref{fig:intro}). 
The learned concepts form interpretable clusters that resemble meaningful manipulation primitives, providing insights into how robots perceive and reason about manipulation tasks.

In summary, our key contributions include: 
(1) a self-supervised framework that extracts structured hierarchical manipulation concepts from unlabeled multi-modal demonstrations, capturing both cross-modal correlations and multi-level temporal abstractions without human annotation; 
(2) an effective policy enhancement approach that integrates these concepts through joint prediction, maintaining compatibility with diverse policy architectures; 
and (3) comprehensive empirical evidence demonstrating significant performance improvements across diverse settings, with analyses revealing how learned concepts enable more robust generalization to novel environments.

\section{Related Work}\label{sec:related work}

\paragraph{Representation Learning in Robotics}
Self-supervised representation learning has emerged as a powerful approach for extracting meaningful skills \citep{liang2024skilldiffuser,liu2025one,rholanguage} from robotic data,
avoiding the need for manual annotation in methods such as
\citep{eisner2022flowbot3d,li2023manipllm,mao2023learning,mo2019partnet}.
Initial efforts explored single-modality approaches for vision-based \citep{bruce2024genie,dasari2023unbiased,seo2023masked,xu2023xskill,ye2024latent} and proprioception-based \citep{lee2024behavior,mete2024quest,pertsch2025fast,shafiullah2022behavior} representation learning.
Recent work integrates multiple modalities,
combining vision with language
\citep{karamcheti2023language,majumdar2023we,nair2022r3m,schmidtlearning,wu2023unleashing}, proprioception with vision
\citep{radosavovic2023robot,wangscaling,yang2023polybot},
even richer
\citep{bonatti2023pact,shah2023mutex,zeng2024learning}. 

These approaches typically focus on cross-modal alignment but often overlook the structured temporal patterns inherent in manipulation tasks.
Parallel developments in temporal representation learning have addressed this challenge through various approaches: time-contrastive learning \citep{li2024decisionnce,ma2023liv,ma2022vip,nair2022r3m,xu2023xskill}, temporal masked auto-encoding \citep{radosavovic2023robot}, and explicit modeling of state transitions across different timescales \citep{kipf2019compile,pertsch2020keyframing,Sharma2020Dynamics-Aware,sharma2023multiresolution}. 
Our work advances the field by simultaneously addressing both multi-modal integration and hierarchical temporal structures, creating representations that naturally align with manipulation sub-goals at varying time horizons while leveraging cross-modal correlational patterns that persist across different objects and contexts.

\paragraph{Concept-Guided Robotic Policies}
Concept-guided approaches enhance robotic policy performance by leveraging intermediate representations that bridge perception and action. These methods generally fall into two categories.
First, two-module frameworks \citep{belkhale2024rt,bu2024towards,liang2024skilldiffuser,wangscaling,ye2024latent,zhouautocgp} employ a dual-model architecture where one component extracts high-level task concepts while another generates the corresponding actions. While effective, these approaches often require specialized architectural designs that limit their applicability across different policy classes.

Second, in contrast, joint prediction approaches \citep{guo2024prediction,jia2024chainofthought,shi2023waypoint,tian2025predictive,yang2022chain} integrate concept guidance by training policies to simultaneously predict both concepts and actions. This creates an implicit information pathway where concept understanding regularizes action generation. Our work adopts this more flexible approach, enabling seamless integration with diverse policy architectures while maintaining the interpretability benefits of explicit concept representations.

\section{Method}
\label{sec:method}

\begin{figure}[!t]
\centering
\includegraphics[width=1.0\linewidth]{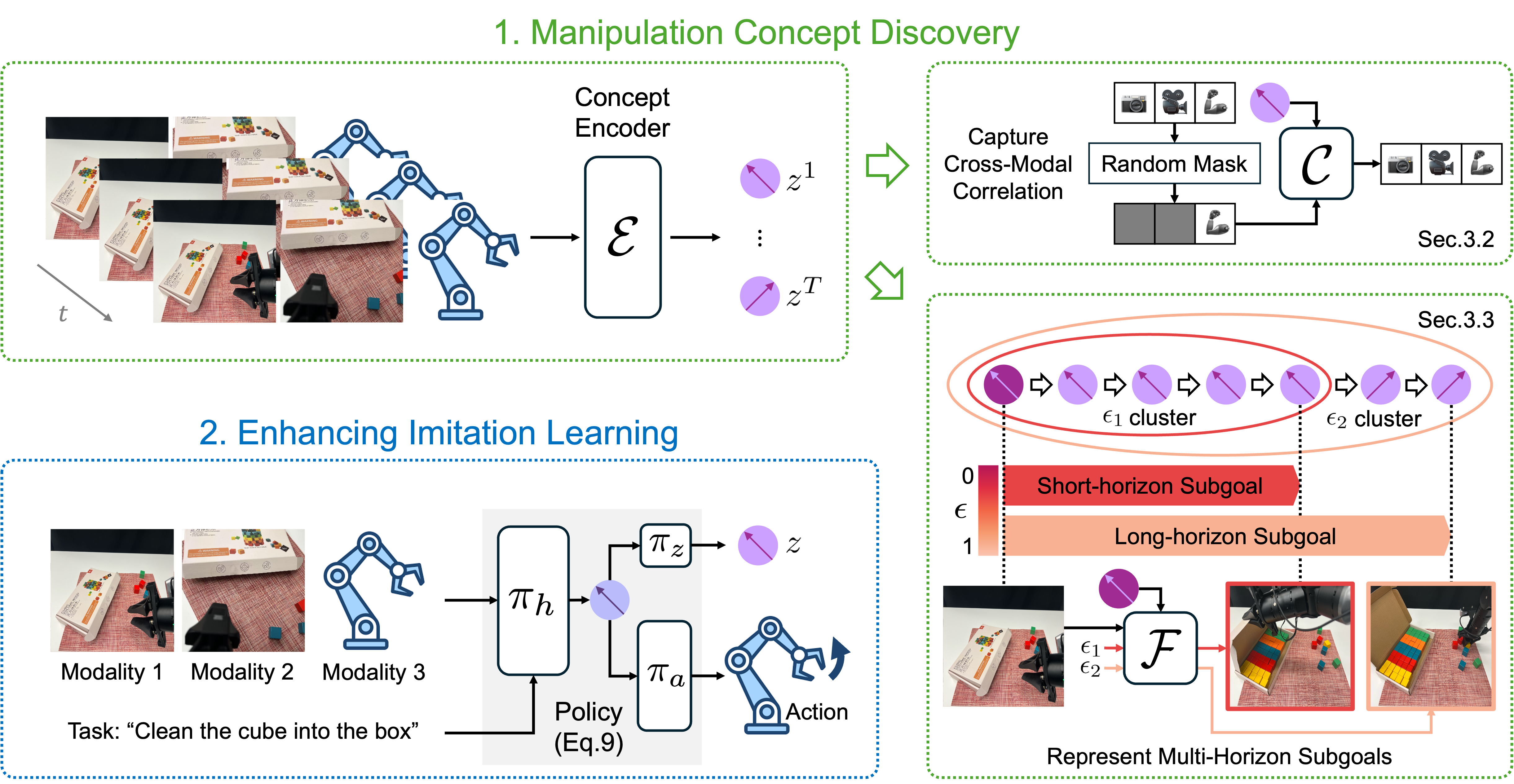}
\vspace{-6mm}
\caption{
\textbf{The proposed self-supervised manipulation concept discovery and policy enhancement.}
\textit{Stage 1:} 
The concept encoder ($\mathcal{E}$) 
processes multi-modal robot demonstrations to extract concept latents. 
These latents are refined through two objectives:
(1) the Cross-Modal Correlation Network ($\mathcal{C}$) employs a mask-and-predict strategy to capture persistent patterns across sensing modalities
(Sec.~\ref{subsec:multi-modal});
(2) the Multi-Horizon Future Predictor ($\mathcal{F}$) enables concept latents to organize hierarchically into multi-horizon sub-goals based on coherence thresholds ($\epsilon$)
(Sec.~\ref{subsec:multi-hierarchy}).
\textit{Stage 2:} 
The learned concepts are integrated into policy learning through a backbone network ($\pi_h$) with concept ($\pi_z$) and action ($\pi_a$) prediction heads, regularizing action generation with structured manipulation knowledge (Eq.~\ref{eq:policy_align_mc}).
}
\label{fig:pipeline}
\vspace{-4mm}
\end{figure}

We aim to encode robotic manipulation demonstrations 
into latent representations that capture task-induced patterns 
in multi-modal sensory-motor data. 
These representations should naturally cluster according to functional sub-goals, providing insights into manipulation objectives and enhancing policy learning. 
We term these clusters \textit{manipulation concepts}---each representing action sequences targeting specific sub-goals---and call the learning process \textit{manipulation concept discovery}.

Our self-supervised approach works without explicit sub-goal annotations, addressing the challenge of capturing meaningful manipulation patterns without labels.
We design objective functions enforcing latent representations that reflect both temporal structure and cross-modal correlations.
Our approach ensures:
(1) integration of modality-specific features while encoding cross-modal correlations that persist across objects and contexts;
(2) hierarchical organization of sub-processes representing sub-goals across temporal horizons and enabling action prediction guided by immediate and long-term objectives.
We validate these concepts through policy performance improvements and multiple analysis methods that demonstrate correspondence with meaningful manipulation primitives.

\subsection{Problem Setup \& Manipulation Concept Encoder}
\label{subsec:mcd}
Given a dataset \(D = \{\tau_i\}_{i=1}^N\) of \(N\) manipulation trajectories,
each \(\tau_i = \{(\mathbf{o}^t_i, a^t_i)\}_{t=1}^{T_i}\) contains observations \(\mathbf{o}^t_i\) and actions \(a^t_i\) at time \(t\).
For \(M\) modalities,
\(\mathbf{o}^t_i = \{o^{1,t}_i, o^{2,t}_i, \dots, o^{M,t}_i\}\),
where \(o^{m,t}_i\) is the observation of the modality \(m\).
We denote \(\mathbf{o}^{S,t}_i = \{o^{m,t}_i \mid m \in S\}\) as observations of modalities in \(S \subseteq [M]=\{1,2,\cdots,M\}\).
We treat observations of both the same sensory modes (e.g., multiple views) and different modes as distinct modalities, as they are functionally different modalities in terms of complementary information.

The \textit{Manipulation Concept Discovery} process assigns latent representation \(z_i^t \in \mathbb{R}^Z\) to each timestep \(t\) of the trajectory \(\tau_i\),
where \( z_i^t \) can be viewed as a noisy sampling of the underlying manipulation concept active at \( t \).
Since these representations cluster based on sub-goals,
we refer to \( z_i^t \) as \textit{manipulation concept latents} or simply \textit{manipulation concepts}.
We use continuous representations for differentiability and to avoid constraints of codebook-based discrete representations (e.g., finite capacity).
To learn \( z_i^t \), we introduce a manipulation concept encoder \(\mathcal{E}\) parameterized by \(\Theta_{\mathcal{E}}\), which maps observation sequence \(\mathbf{o}_i=\{\mathbf{o}_i^t\}_{t=1}^{T_i}\) from trajectory \(\tau_i\) to concept sequence \(\mathbf{z}_i = \{z_i^t\}_{t=1}^{T_i}\):
\begin{equation}
\mathbf{z}_i \leftarrow \mathcal{E}\left(\mathbf{o}_i;\Theta_{\mathcal{E}}\right)
\label{eq:concept_encoder}
\end{equation}
We implement \(\mathcal{E}\) 
using a transformer to encode temporal dependencies (details in Sec.~\ref {subsec:id_MCD}). 
Next, we elaborate on the training strategies optimizing cross-modal and multi-horizon temporal correlation metrics (Sec.~\ref{subsec:multi-modal} and~\ref{subsec:multi-hierarchy}).

\subsection{Capturing Multi-Modal Correlations}
\label{subsec:multi-modal}

To enhance the utility of multi-modal information,
we propose that manipulation concepts should capture \textit{cross-modal correlations} rather than simply aggregating features from different modalities
(e.g., concatenating multi-modal signals \citep{chi2023diffusion,zhaolearning}).
Physiological evidence suggests that concept formation often occurs when correlations across sensory modalities are high
\citep{axmacher2006memory,fell2011role,melloni2007synchronization,singer2011consciousness,womelsdorf2007role}.
These correlations remain consistent across scenarios involving the same concept,
facilitating generalization. For instance, in container opening tasks, the correlated patterns between visual lid rotation, characteristic force feedback, and audio cues persist across different container types, enabling the transfer of the ``opening'' concept despite variations in object appearances.

To learn manipulation concepts that capture cross-modal correlations,
we propose maximizing mutual information---a metric capable of modeling diverse correlations---between observations from different modalities, conditioned on the associated manipulation concept.
Specifically, we maximize the conditional mutual information over bipartitions of modality observations:
\begin{equation}
\begin{aligned}
\max_\mathbf{Z}\sum_{S\subsetneq\left[M\right],S\neq\emptyset}
\mathbb{I}
\left(
\mathbf{O}_{S} : \mathbf{O}_{\left[M\right]\setminus S} \mid \mathbf{Z}
\right),
\end{aligned}
\label{eq:mi_and_moc}
\end{equation}
where \(\mathbf{O}_{S}\) are observations from a subset of modalities, \(\mathbf{O}_{\left[M\right]\setminus S}\) are observations from remaining modalities, and \(\mathbf{Z}\) is the manipulation concept.
We implement Eq.~\ref{eq:mi_and_moc} using a computationally efficient self-supervised \textit{mask-and-predict} approach that stochastically samples bipartitions during training.
This ensures scalability despite exponentially increasing bipartition numbers while integrating cross-modal correlation learning with multi-modal information compression.

Specifically, a Cross-Modal Correlation Network 
\(\mathcal{C}\) (CMCN) 
with parameters \(\Theta_c\) reconstructs full-modality observations from partial observations guided by manipulation concepts.
During training, 
we mask observations from a random subset \(S\) of modalities 
and reconstruct all observations 
\(\mathbf{o}^t_i\) using the unmasked subset \(\mathbf{o}^{[M]\setminus S,t}_i\) 
and concept \(z^t_i\):
\begin{equation}
\begin{aligned}
&\mathcal{L}_{\text{mm}}\left(t,\mathbf{\tau}_i\right)
=\mathbb{E}_{S}\left\| \mathcal{C}\left(\textbf{o}^{[M]\setminus S,t}_i, z^t_i;\Theta_c\right) - \textbf{o}^t_i \right\|,
\end{aligned}
\label{eq:loss_mi_and_moc}
\end{equation}
where \(S \sim \text{U}\left(2^{[M]} \setminus \{\emptyset\}\right)\) is a uniformly sampled non-empty subset of modality indices.
By predicting full observations from partial inputs,
we maximize the conditional mutual information in Eq.~\ref{eq:mi_and_moc},
forcing manipulation concepts \(z^t_i\) to capture cross-modal correlations.
Additionally, when all modalities are masked,
reconstruction solely from \(z^t_i\) ensures these representations compress and preserve essential multi-modal information (please see Sec.~\ref{subsec:id_MCD} for more details).

\subsection{Representing Multi-Horizon Sub-Goals}
\label{subsec:multi-hierarchy}

To complete tasks with hierarchical structures,
manipulation concepts must encode multi-horizon sub-goal information.
Physiological evidence shows human actions are hierarchically organized \citep{grafton2007evidence,murray2014hierarchy},
with coarse-grained goals defining overall tasks and fine-grained goals informing immediate actions.
These multi-horizon sub-goals link ultimate goals with low-level actions,
enabling smooth transitions while enhancing robustness.

We aim to make manipulation concepts organized to encode sub-processes across multiple temporal horizons without explicit annotations.
Since concepts cluster by sub-goals,
hierarchical sub-goals can emerge from these clusters at varying temporal scales.
We propose that the temporal extent of a sub-process is determined by concept latent coherence within clusters,
yielding a natural spectrum from short-horizon to long-horizon sub-goals.
Specifically, 
given manipulation concept latents 
\(\mathbf{z}_i = \{z_i^t\}_{t=1}^{T_i}\) from trajectory \(\tau_i\),
we quantify their similarities 
using spherical distance:
\(\mathrm{dist}(z, u) = \frac{1}{\pi} \arccos{\left\langle \frac{z}{\lVert z \rVert_2}, \frac{u}{\lVert u \rVert_2} \right\rangle}\).
Concepts belong to the same sub-process if their distance falls below a coherence threshold \(\epsilon \in [0,1]\).
More explicitly, sub-processes are derived as:
\begin{equation}
\begin{aligned}
    &\mathrm{h}\left(\mathbf{z}_i;\epsilon\right) = \big\{[g_k, g_{k+1}) \mid k = 1,2,\cdots,K(\mathbf{z}_i;\epsilon)\big\},\\
    & \text{where }g_1=1,\hspace{2mm}
    g_{k+1}=\max_g
    \big\{ g \mid g\in(g_k,T_i+1]\cap\mathbb{N}^+ \wedge \forall t,t' \in [g_k,g),\mathrm{dist}(z^t_i, z^{t'}_i) < \epsilon \big\},
\end{aligned}
\label{eq:t_structure}
\end{equation}
where \(K(\mathbf{z}_i;\epsilon)\) is the number of clusters determined by \(\epsilon\), and increasing \(\epsilon\) yields sub-processes spanning from short-horizon to long-horizon. 
Please see Alg.~\ref{alg:e_cluster} for more details.

Furthermore, 
we propose learning objectives 
to ensure multi-horizon sub-processes from Eq.~\ref{eq:t_structure} align with meaningful sub-goal completion processes.
Specifically, 
the manipulation concept guiding each sub-process should be informative about the state achieved upon sub-task completion \citep{blackzero,lynch2020learning,zhen3d}.
For all coherence thresholds \(\epsilon\),
current observation \(\mathbf{O}\) and its associated concept \(\mathbf{Z}\) should be informative of the terminal observation \(\mathbf{O}^{\mathbf{goal}(\epsilon)}\),
characterized by minimizing the following conditional entropy:
\begin{equation}
\forall\epsilon, \min_\mathbf{Z} \mathbb{H}\Big(\mathbf{O}^{\mathbf{goal}(\epsilon)} \mid \mathbf{O}, \mathbf{Z}\Big),
\label{eq:predictable goal}
\end{equation}
To implement Eq.~\ref{eq:predictable goal},
we train a Multi-Horizon Future Predictor \(\mathcal{F}\) (MHFP) to hallucinate terminal observations of different sub-processes.
For time step \(t\) in trajectory \(\tau_i\),
the terminal observation is determined by the ending time step of the interval containing \(t\):  
\begin{equation}
\mathrm{g}(t;\mathbf{z}_i,\epsilon) = \min\{T_i, g_{k+1}\}, \, \text{where }t \in [g_k, g_{k+1}) \in \mathrm{h}(\mathbf{z}_i;\epsilon),
\label{eq:t_terminal}
\end{equation}
During training, 
the network \(\mathcal{F}\), parameterized by \(\Theta_f\),
predicts this terminal observation based on
current observation \(\mathbf{o}^t_i\),
manipulation concept \(z^t_i\),
and coherence threshold \(\epsilon\): 
\begin{equation}
\begin{aligned}
\mathcal{L}_{\text{mh}}\left(t, \tau_i\right)
= \mathbb{E}_{\epsilon} \left\| \mathcal{F}\left(\mathbf{o}^t_i, z^t_i, \epsilon; \Theta_f\right) - \mathbf{o}^{\mathrm{g}(t;\mathbf{z}_i,\epsilon)}_i \right\|,
\end{aligned}
\label{eq:loss_predictable goal}
\end{equation}
where $\epsilon \sim \mathrm{U}([0,1])$ is sampled uniformly per iteration to improve efficiency by avoiding training over all $\epsilon$ values.
This training process iteratively improves both latents and sub-process derivation:
we compute manipulation concepts using the encoder (Eq.~\ref{eq:concept_encoder}),
determine sub-process boundaries,
then update all networks,
including \(\mathcal{F}\) and the concept encoder.
This improves future observation prediction and concept latents,
which in turn refines sub-process derivation.
By minimizing Eq.~\ref{eq:loss_predictable goal},
\(z^t_i\) is ensured to encode multi-horizon sub-goal information,
indicating hierarchical transitions to terminal states under various \(\epsilon\) while 
adjusting sub-processes by shaping concept latents for terminal state predictability.
More details can be found in Sec.~\ref{subsec:id_MCD}.

\textit{\textbf{Final Objective for Manipulation Concept Discovery.}}
We jointly optimize the multi-modal correlation objective (Eq.~\ref{eq:loss_mi_and_moc}) and multi-horizon sub-goal prediction objective (Eq.~\ref{eq:loss_predictable goal}) 
to ensure manipulation concepts 
generated by the encoder \(\mathcal{E}\) (Eq.~\ref{eq:concept_encoder}) satisfy both key properties:
\begin{equation}
\mathcal{L}_\text{z}\left(t, \tau_i\right) = \lambda_\text{mm}\mathcal{L}_\text{mm}\left(t, \tau_i\right) + \lambda_\text{mh}\mathcal{L}_\text{mh}\left(t, \tau_i\right),
\label{eq:joint_mcd_loss}
\end{equation}  
where \(\lambda_{\text{mm}}, \lambda_{\text{mh}} > 0\) balance the two loss terms.

\subsection{Enhancing Imitation Learning with Manipulation Concepts}\label{subsec:EIL}
After learning manipulation concepts through our self-supervised framework,
we address how these concepts enhance policy learning.
Unlike previous approaches that learn task-specific policies directly from demonstrations \citep{ding2023task,li2023manipllm},
we propose to leverage the learned manipulation concepts as an informative representation that bridges low-level actions and high-level goals.

Specifically, with manipulation concepts 
\(\mathbf{z}_i\) generated by encoder \(\mathcal{E}\),
we augment imitation learning by training policies to predict both ground-truth actions and corresponding concepts \citep{jia2024chain,yang2022chain,zhao2025cot}.
This approach uses concept prediction 
as a regularization that guides the policy to encode conceptual understanding alongside action planning:
\begin{equation}
\begin{aligned}
& h_i^t = \pi_h\left(\mathbf{o}^t_i, \ell_i; \Theta_\pi^h\right),
\quad\hat{z}^t_i = \pi_z\left(h_i^t; \Theta_\pi^z\right),
\quad\hat{a}^t_i = \pi_a\left(h_i^t; \Theta_\pi^a\right), \\
& \mathcal{L}_{\pi}(t, \tau_i, \ell_i) = \lVert\hat{a}^t_i - a^t_i\rVert + \lambda_{\text{mc}}\lVert\hat{z}^t_i - z^t_i\rVert.
\end{aligned}
\label{eq:policy_align_mc}
\end{equation}
The policy consists of:
(1) A backbone \(\pi_h\) processing task descriptions \(\ell_i\) and observations \(\mathbf{o}^t_i\) to produce a shared representation \(h_i^t\);
(2) A concept predictor \(\pi_z\) mapping \(h_i^t\) to predicted concepts \(\hat{z}^t_i\);
and (3) An action decoder \(\pi_a\) mapping \(h_i^t\) to predicted actions \(\hat{a}^t_i\).
This joint objective enforces the policy to leverage concept information encoded within \(h_i^t\) while predicting actions.
Even though concepts are learned task-agnostically for generalization,
the policy receives task descriptions in a multi-task setting,
serving as a mechanism to learn the reuse of concepts.
The learning objective balances action and concept prediction using \(\lambda_{\text{mc}} > 0\). More details are provided in Sec.~\ref{subsec:id_EIL}.

\section{Experiments}
\label{sec:experiments}

We evaluate our manipulation concept discovery approach through experiments addressing four key questions:
(1) Do learned concepts enhance policy performance on tasks used for concept discovery, validating our strategies for encoding cross-modal correlations (Sec.~\ref{subsec:multi-modal}) and multi-horizon sub-goals (Sec.~\ref{subsec:multi-hierarchy})?
(2) Can concepts learned from one task set transfer effectively to different tasks sharing underlying manipulation patterns?
(3) Does our concept discovery mechanism generalize to novel tasks with decreased overlap in manipulation patterns?
(4) What interpretable properties emerge in the learned concepts that explain their effectiveness for robotic manipulation?
Through these investigations, we demonstrate both the immediate benefits of our approach for imitation learning and its broader applicability for transfer learning and generalization in manipulation tasks.

\subsection{Experimental Setup}
\label{subsec:exp_setups}

\paragraph{Dataset and Environment}
Sec.~\ref{subsec:main_results} and \ref{subsec:direct_concept} conduct experiments using the \textbf{LIBERO} benchmark \citep{LIBERO}, a comprehensive platform for robotic learning built on Robosuite \citep{robosuite2020}.
We utilize three distinct task sets:
\begin{itemize}[left=0em, itemsep=0pt, parsep=0pt]
    \item \textbf{LIBERO-90}: A diverse collection of 90 manipulation tasks serving as our primary training domain for concept discovery and initial policy learning.
    \item \textbf{LIBERO-LONG}: 10 novel long-horizon tasks, each composed of two LIBERO-90 tasks in sequence, designed to evaluate transfer to more complex task structures.
    \item \textbf{LIBERO-GOAL}: 10 tasks in an entirely novel environment unseen during concept discovery, used to evaluate the generalization of learned concepts to unfamiliar contexts.
\end{itemize}
Each task includes a natural language description and 50 expert demonstrations. For multi-modal observations, we use:
\textit{Agentview vision}: 128×128 RGB third-person camera capturing the entire environment;
\textit{Eye-in-hand vision}: 128×128 RGB gripper-mounted camera;
\textit{Proprioceptive state}: 9D vector encoding gripper position, rotation, and physical states.

\paragraph{Manipulation Concept Discovery Methods}
We compare our approach with several state-of-the-art concept discovery baselines (implementation details in Sec.~\ref{subsec:id_baselines}):
\begin{itemize}[left=0em, itemsep=0pt, parsep=0pt]
    \item \textbf{InfoCon} \citep{liuinfocon}: A VQ-VAE type of method for single-hierarchy concept discovery.
    \item \textbf{XSkill} \citep{xu2023xskill}: Contrastive learning for manipulation skill extraction from demonstration videos.
    \item \textbf{DecisionNCE} \citep{li2024decisionnce}: Learns reward-relevant representations from demonstrations with language annotations, evaluated in two variants: using task instructions (DecisionNCE-task) and using elementary action labels (DecisionNCE-motion).
    \item \textbf{RPT} \citep{radosavovic2023robot}: Temporally and modality-masked autoencoder for multi-modal sequence modeling.
    \item \textbf{All}: A simplified variant of our approach that predicts all modalities from concepts without modeling cross-modal correlations.
    \item \textbf{Next}: Predicts adjacent time-step observations, a common approach adopted in \citep{bruce2024genie,ye2024latent}.
    \item \textbf{CLIP} \citep{clip}: Language-aligned visual features from a pretrained foundation model.
    \item \textbf{DINOv2} \citep{oquab2023dinov2}: Self-supervised visual representations without temporal modeling.
    \item \textbf{Plain}: Standard imitation learning without manipulation concepts.
\end{itemize}

\paragraph{Policies for Concept-Enhanced Imitation Learning}
To evaluate the effectiveness of our discovered manipulation concepts, we integrate them into two established imitation learning frameworks using the joint prediction approach described in Sec.~\ref{subsec:EIL}:

\begin{itemize}[left=0em, itemsep=0pt, parsep=0pt]
    \item \textbf{ACT} \citep{zhaolearning}: A transformer-based conditional variational autoencoder that predicts action chunks.
    \item \textbf{Diffusion Policy (DP)} \citep{chi2023diffusion}: A 1D convolutional UNet that generates actions through denoising.
\end{itemize}

For both policy architectures, we add the concept prediction head (\(\pi_z\) in Eq.~\ref{eq:policy_align_mc}) to predict manipulation concepts from the shared concept-aware representations. 
Implementation details appear in Sec.~\ref{subsec:id_EIL}. 
All experiments are reported with results aggregated across 4 random seeds.

\subsection{Evaluating Policy Performance with Learned Manipulation Concepts}

\label{subsec:main_results}

\begin{table}[!t]
\centering
\small
\caption{
\textbf{Evaluation of manipulation concept discovery methods across different task settings.}
Success rates (\%) of ACT and Diffusion Policy (DP) models when enhanced with manipulation concepts from various discovery methods. 
All concept encoders were trained only on LIBERO-90, and evaluated on: original tasks (\textbf{\textit{L90-90}}), novel long-horizon compositions (\textbf{\textit{L90-L}}), and entirely new environments (\textbf{\textit{L90-G}}). Values in parentheses show standard deviations across 4 seeds. \textbf{Bold} and \underline{underlined} values indicate best and second-best results.
}
\begin{tabular}{
w{c}{0.8cm}w{c}{0.7cm}w{c}{0.7cm}w{c}{0.7cm}w{c}{0.7cm}w{c}{0.7cm}w{c}{0.7cm}w{c}{0.7cm}w{c}{0.7cm}w{c}{0.9cm}w{c}{0.7cm}w{c}{0.7cm}
}
\toprule
\multirow{2}{*}{\textit{\textbf{L90-90}}} & \multirow{2}{*}{InfoCon} & \multirow{2}{*}{XSkill} & \multirow{2}{*}{RPT} & \multirow{2}{*}{All} & \multirow{2}{*}{Next} & \multirow{2}{*}{CLIP} & \multirow{2}{*}{DINOv2} & \multicolumn{2}{c}{DecisionNCE} & \multirow{2}{*}{Plain} & \multirow{2}{*}{\textbf{Ours}} \\
& & & & & & & & task & motion & & \\
\midrule
\multirow{2}{*}{ACT} & 66.5 & \underline{73.4} & 68.8 & 64.1 & 68.0 & 63.8 & 71.9 & 69.0 & 66.8 & 46.6 & \textbf{74.8} \\
                     & (0.8) & (0.8) & (0.8) & (2.0) & (0.4) & (0.5) & (0.3) & (0.1) & (0.8) & (1.9) & (0.8) \\
\midrule
\multirow{2}{*}{DP}  & 78.2 & \underline{87.7} & 84.3 & 81.5 & 82.6 & 80.7 & 79.4 & 75.7 & 82.7 & 75.1 & \textbf{89.6} \\
                     & (0.6) & (0.6) & (0.1) & (0.5) & (0.1) & (0.9) & (0.1) & (0.8) & (0.6) & (0.6) & (0.6) \\
\midrule
\midrule
\multirow{2}{*}{\textit{\textbf{L90-L}}} & \multirow{2}{*}{InfoCon} & \multirow{2}{*}{XSkill} & \multirow{2}{*}{RPT} & \multirow{2}{*}{All} & \multirow{2}{*}{Next} & \multirow{2}{*}{CLIP} & \multirow{2}{*}{DINOv2} & \multicolumn{2}{c}{DecisionNCE} & \multirow{2}{*}{Plain} & \multirow{2}{*}{\textbf{Ours}} \\
& & & & & & & & task & motion & & \\
\midrule
\multirow{2}{*}{ACT} & 55.5 & 55.0 & \underline{59.0} & 55.5 & 55.0 & 51.0 & 55.0 & 53.0 & 49.3 & 54.0 & \textbf{63.0} \\
                     & (0.9) & (1.0) & (1.0) & (0.9) & (1.0) & (1.0) & (1.0) & (1.0) & (0.9) & (0.9) & (1.0) \\
\midrule
\multirow{2}{*}{DP} & 75.0 & 73.0 & 61.3 & 79.3 & \underline{83.0} & 67.0 & 63.0 & 58.7 & 52.7 & 34.1 & \textbf{89.0} \\
                    & (1.0) & (1.0) & (0.9) & (0.9) & (1.0) & (1.0) & (1.0) & (0.9) & (0.9) & (1.1) & (1.0) \\

\midrule
\midrule
\multirow{2}{*}{\textit{\textbf{L90-G}}} & \multirow{2}{*}{InfoCon} & \multirow{2}{*}{XSkill} & \multirow{2}{*}{RPT} & \multirow{2}{*}{All} & \multirow{2}{*}{Next} & \multirow{2}{*}{CLIP} & \multirow{2}{*}{DINOv2} & \multicolumn{2}{c}{DecisionNCE} & \multirow{2}{*}{Plain} & \multirow{2}{*}{\textbf{Ours}} \\
& & & & & & & & task & motion & &  \\
\midrule
\multirow{2}{*}{ACT} & 67.0 & 77.0 & 75.0 & 69.0 & 71.0 & 77.0 & \underline{77.3} & 70.0 & 75.0 & 57.0 & \textbf{81.0} \\
                     & (1.0) & (1.0) & (1.0) & (1.0) & (1.0) & (1.0) & (0.9) & (0.9) & (0.5) & (1.0) & (1.0) \\
\midrule
\multirow{2}{*}{DP} & 92.7 & \underline{93.0} & 91.5 & 91.0 & 91.3 & 92.0 & 91.0 & 92.0 & 93.0 & 90.7 & \textbf{95.7} \\
                    & (0.9) & (1.0) & (0.9) & (1.0) & (0.9) & (0.9) & (0.7) & (0.8) & (1.0) & (0.9) & (0.7) \\
\bottomrule
\end{tabular}
\label{tab:main-results}
\vspace{-3mm}
\end{table}

\begin{itemize}[left=0em, itemsep=0pt, parsep=0pt]
\item \textbf{Performance on Original Training Tasks}\hspace{2mm}
We first evaluate our concept discovery method on the same tasks used for concept training. 
As shown in the \textbf{\textit{L90-90}} results (Tab.~\ref{tab:main-results}), our approach consistently outperforms all baselines with both policy architectures. 
The performance gap between our method and \textit{Next}/\textit{InfoCon} demonstrates the importance of multi-hierarchical sub-goal modeling, while improvements over \textit{All} highlight the value of explicitly capturing cross-modal correlations. 
Our method also surpasses \textit{DecisionNCE} variants despite not requiring language supervision, validating the effectiveness of our self-supervised objectives.
\item \textbf{Transfer to Long-Horizon Tasks}\hspace{2mm}
To evaluate concept transferability to more complex compositions, we apply concept encoders trained on LIBERO-90 \textit{directly} to LIBERO-LONG demonstrations featuring novel long-horizon tasks. The \textbf{\textit{L90-L}} results show our method maintains its performance advantage in this challenging transfer setting. This demonstrates that our approach learns manipulation concepts that effectively decompose hierarchical tasks, enabling policies to better handle novel complex task compositions requiring sequential execution of multiple sub-goals.
\item \textbf{Generalization to Novel Environments}\hspace{2mm}
We further test generalization by applying concept encoders trained on LIBERO-90 \textit{directly} to LIBERO-GOAL demonstrations featuring unseen environments and tasks. The \textbf{\textit{L90-G}} results show our method continues to outperform all baselines in this challenging scenario. This indicates our approach discovers fundamental manipulation primitives that transfer effectively across environmental variations.
\item \textbf{Impact of Multi-Modal Observations}\hspace{2mm}
Our ablation study (Tab.~\ref{tab:modal_ablation}) shows that performance consistently improves as more modalities are incorporated. The most significant drops occur when removing proprioceptive information, highlighting its importance for grounding visual observations with physical interaction states and confirming the value of our cross-modal correlation approach.
\end{itemize}

\begin{table}[!t]
\caption{\textbf{Impact of modality combinations on concept discovery performance.}
Success rates (\%) of ACT and DP policies using manipulation concepts discovered with different input modality combinations. All models were trained and evaluated on LIBERO-90, with specific modalities excluded (marked with ``--''). 
A: agentview vision, H: eye-in-hand vision, P: proprioceptive state.}
\centering
\begin{tabular}{cccccccc}
\toprule
 & \textbf{Ours} & -- H P & A -- P & A H -- & -- -- P & -- H -- & A -- -- \\
\midrule
ACT & 74.8±0.8 & 70.5±1.8 & 71.3±0.3 & 70.1±1.2 & 67.5±0.8 & 68.7±0.6 & 69.4±0.4 \\
\midrule
DP & 89.6±0.6 & 85.8±0.2 & 85.6±0.3 & 84.3±0.5 & 84.8±0.1 & 83.7±0.1 & 85.3±0.5 \\
\bottomrule
\end{tabular}
\label{tab:modal_ablation}
\vspace{-3mm}
\end{table}

\subsection{Analyzing Manipulation Concept Properties}
\label{subsec:direct_concept}

\paragraph{Enhanced Cross-Modal Correlation}
\begin{wrapfigure}{r}{0.37\textwidth}
  \vspace{-4mm}
  \centering
  \captionof{table}{
  \textbf{Conditional mutual information between modality pairs.}
  Values conditioned on concept latents from our method versus the \textbf{All} baseline that does {\it not} model cross-modal correlations. A: agentview, H: eye-in-hand vision, P: proprioception.}
  \vspace{-2mm}
    \begin{tabular}{cccc}
        \toprule
         & Ours & All \\
        \midrule
        \(\mathbb{I}\left(\mathbf{o}_{\text{H}}:\mathbf{o}_{\text{A}}\mid \mathbf{z}\right)\) & 3.7999 & 2.0080 \\
        \midrule
        \(\mathbb{I}\left(\mathbf{o}_{\text{P}}:\mathbf{o}_{\text{A}}\mid\mathbf{z}\right)\) & 4.8319 & 3.1312 \\
        \midrule
        \(\mathbb{I}\left(\mathbf{o}_{\text{P}}:\mathbf{o}_{\text{H}}\mid \mathbf{z}\right)\)  & 4.8255 & 3.1322 \\
        \bottomrule
    \end{tabular}
    \label{tab:mi}
    \vspace{-5mm}
\end{wrapfigure}

\begin{wrapfigure}{r}{0.5\textwidth}
    \vspace{0mm}
    \centering
    \includegraphics[width=0.5\textwidth]{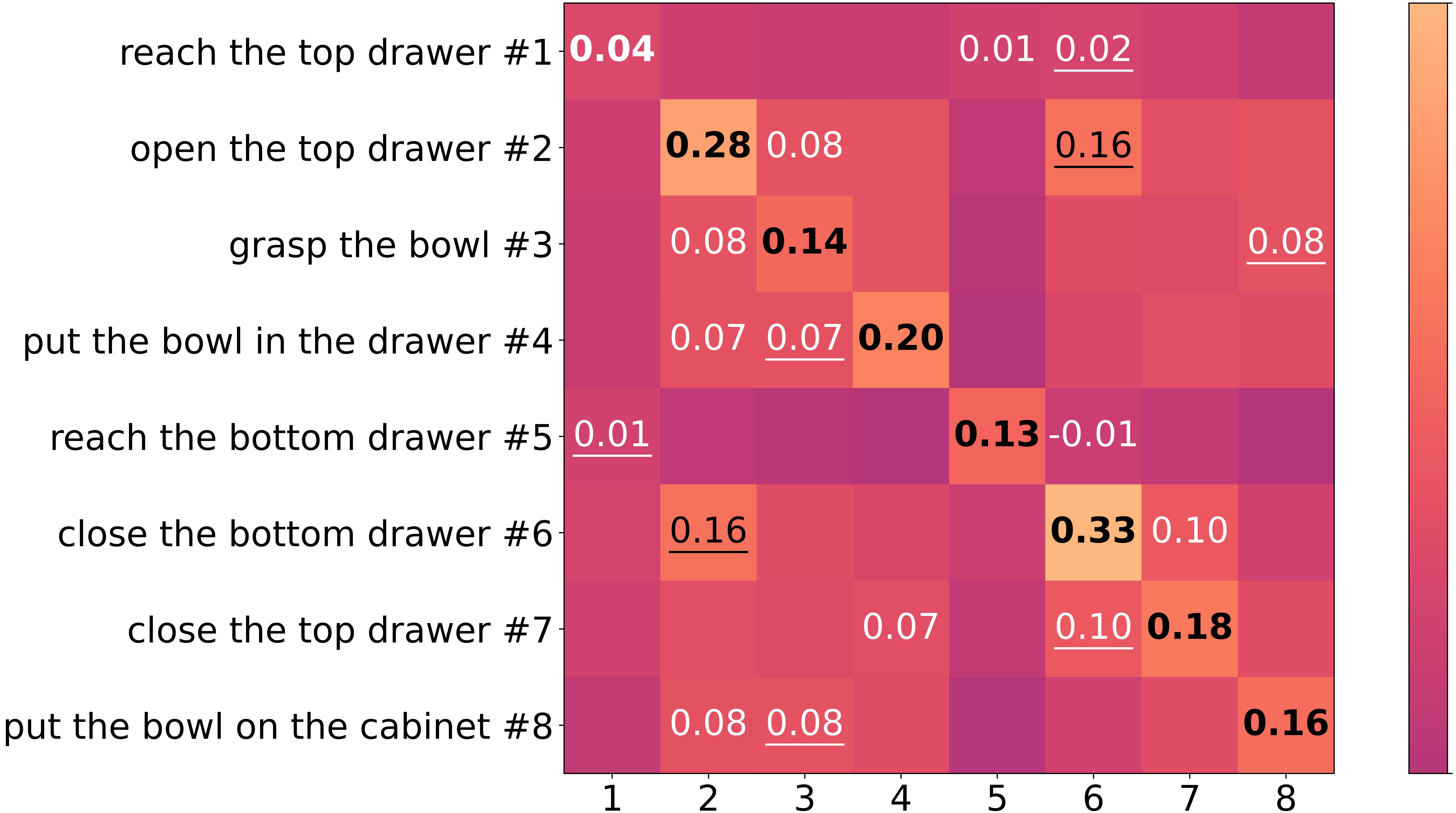}
    \captionof{figure}{
    \textbf{Semantic alignment of learned concepts.}
    Cosine similarity between concept latents grouped by human-defined sub-goals. Diagonal patterns demonstrate that our approach discovers concepts that exhibit clustering patterns corresponding to meaningful manipulation primitives.}
    \label{fig:semantic_example}
    \vspace{-6mm}
\end{wrapfigure}

To verify our Cross-Modal Correlation Network's effectiveness (Sec.~\ref{subsec:multi-modal}), we measure mutual information between modalities conditioned on concept latents (Sec.~\ref{subsec:id_MI}).
Tab.~\ref{tab:mi} shows that our approach achieves higher conditional mutual information than the \textbf{All} baseline. This confirms that our mask-and-predict strategy enables the concept encoder to capture persistent cross-modal patterns that generalize across different objects and contexts, providing a robust representational basis for policies.
\vspace{-3mm}
\paragraph{Alignment with Semantic Sub-Goals}
We evaluate whether our concepts align with human-understandable semantics by grouping latents from different demonstrations based on human-identified sub-goals and computing similarities between these groupings:
\begin{equation}
\langle C_i, C_j \rangle = \frac{1}{|C_i||C_j|} \sum_{z_i \in C_i} \sum_{z_j \in C_j} \left\langle \frac{z_i}{\lVert z_i \rVert_2}, \frac{z_j}{\lVert z_j \rVert_2} \right\rangle,
\label{eq:class_sim}
\end{equation}
where \( C_i \), \(C_j\) represent 
human-identified sub-goal categories, 
and \( z_{i} \), \( z_j \) are latents within each category (details in Sec.~\ref{subsec:semantic study}).
As shown in Fig.~\ref{fig:semantic_example}, similarity matrices consistently show the highest values along the diagonal, demonstrating that our approach discovers concepts that exhibit clustering patterns corresponding to meaningful manipulation primitives.

\begin{figure}[!t]
\centering
\includegraphics[width=1.0\linewidth]{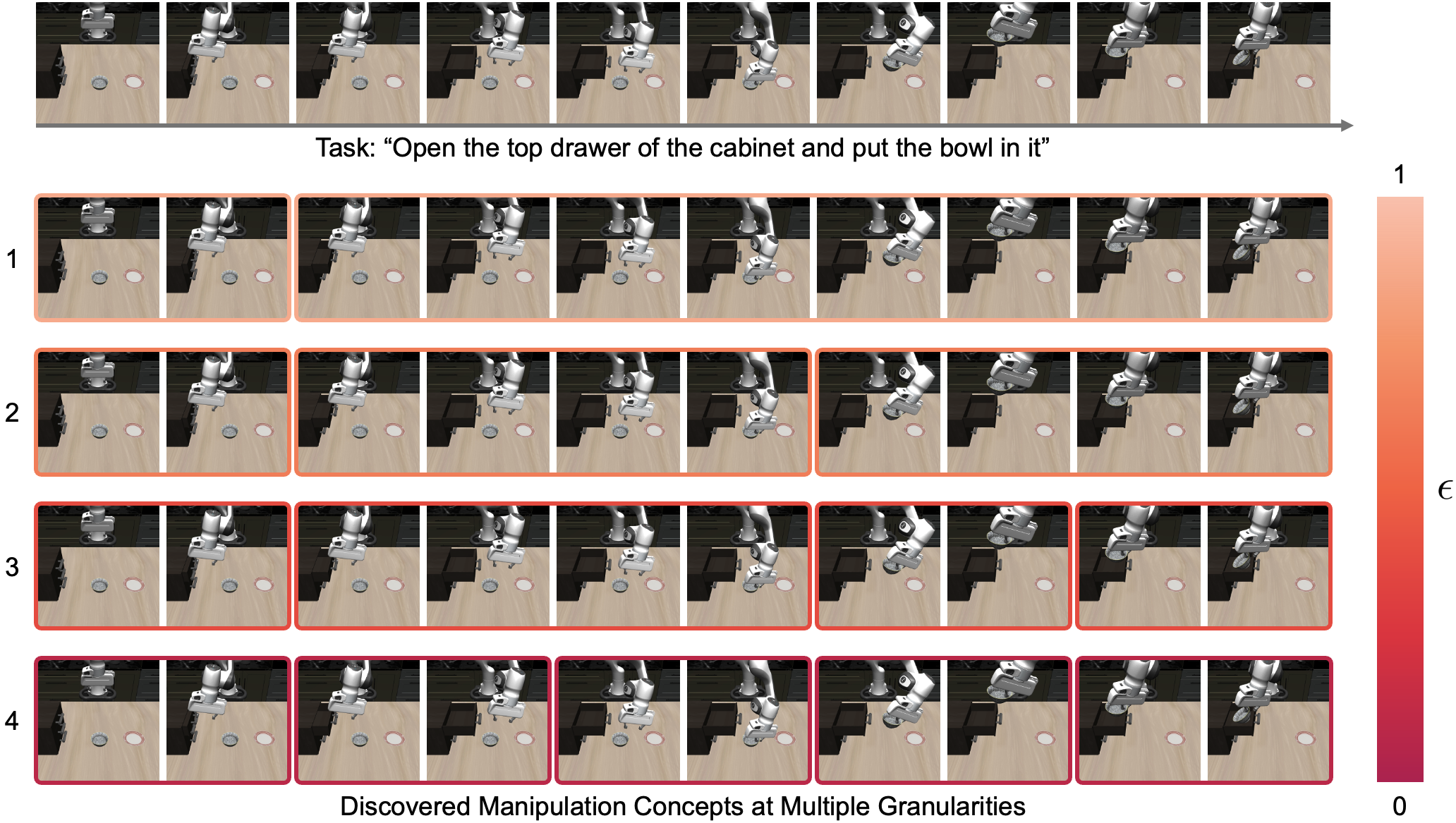}
\vspace{-5mm}
\caption{
\textbf{Multi-granular task decomposition through concept latent clustering.}
Visualization of sub-processes derived by clustering manipulation concept latents at different coherence thresholds ($\epsilon$) for the task ``open the top drawer and put the bowl in it.'' 
Higher $\epsilon$ values (top rows) produce coarser decompositions, while lower values (bottom rows) yield finer-grained segmentation. 
The emergent sub-processes naturally align with semantic task components, for example, the third segment in row 2 corresponds to ``put bowl in drawer,'' while the second segment in row 4 corresponds to ``pull drawer open.'' 
This demonstrates our method's ability to discover hierarchical, human-interpretable task structures without explicit supervision.}
\label{fig:vis_hierarchy_example}
\vspace{-5mm}
\end{figure}

\vspace{-3mm}
\paragraph{Multi-Level Hierarchical Structure}
Varying the coherence threshold $\epsilon$ in Eq.~\ref{eq:t_structure} reveals the hierarchical organization of our learned concepts. Fig.~\ref{fig:vis_hierarchy_example} (and Sec.~\ref{subsec:multi horizon subgoals}) shows larger $\epsilon$ values identify coarse-grained phases,
while smaller values capture fine-grained actions.
This emergent hierarchy enables policies to simultaneously reason about immediate actions and longer-term goals without explicit hierarchical supervision, contributing to improved performance on complex sequential tasks that require coordinated execution across multiple temporal scales.

\subsection{Real-World Validation}
\label{subsec:real_world}
\begin{figure}[!t]
\centering
\includegraphics[width=1.0\linewidth]{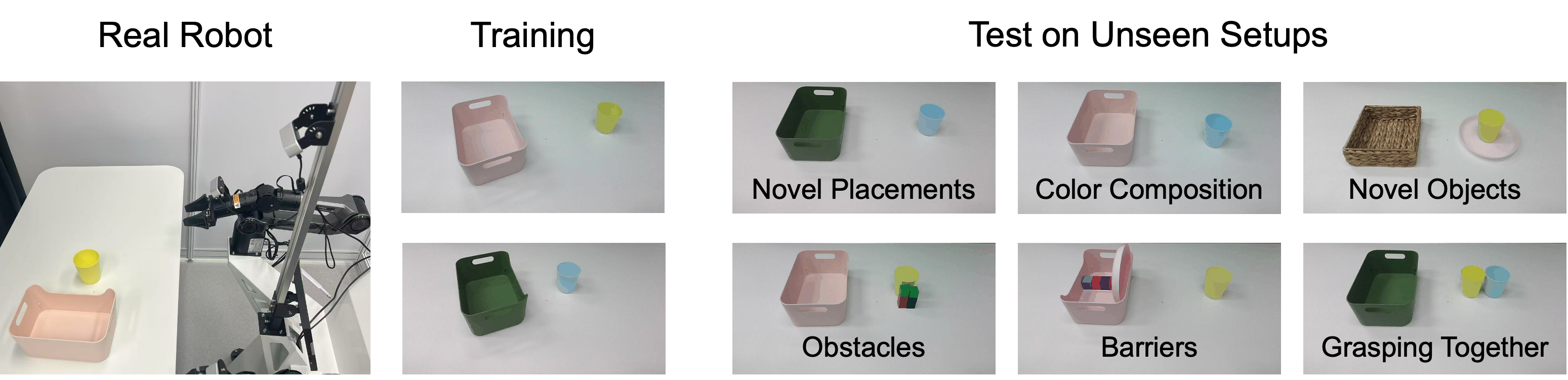}
\caption{
\textbf{Real-world generalization evaluation with Mobile ALOHA robot.}
Left: Mobile ALOHA robot setup for cup cleaning tasks. Center: Training conditions with simple, consistent cup-container color pairings. Right: Six test variations with increasing complexity: novel placements, altered color combinations, unfamiliar objects, external obstacles, internal barriers, and simultaneous grasping of multiple cups. These variations test the policy's ability to generalize beyond training conditions by systematically introducing new challenges.}
\label{fig:real_exp_setup_main}
\vspace{-5mm}
\end{figure}

\paragraph{Real-World Generalization Study}
To study generalization capabilities,
we deploy concept-enhanced policies on a Mobile ALOHA robot \citep{fu2024mobile} in ``cleaning cup'' tasks (Fig.~\ref{fig:real_exp_setup_main}). 
Training data includes only simple container arrangements with consistent color pairings.
We test on six increasingly challenging variations:
\textbf{(1) Novel Placements}: Cups and containers in unseen arrangements;
\textbf{(2) Color Composition}: Altered cup-container color pairings;
\textbf{(3) Novel Objects}: Entirely unseen containers, cups, and plates;
\textbf{(4) Obstacles}: Objects between the robot and the cups obstructing vision;
\textbf{(5) Barriers}: Internal dividers within containers impeding placement;
\textbf{(6) Grasping Together}: Two adjacent cups requiring simultaneous grasp.

As shown in Tab.~\ref{tab:real-exp-result}, policies enhanced with our manipulation concepts consistently outperform baselines across all scenarios, with advantages in challenging conditions.
We suggest that the following two mechanisms behind learned manipulation concepts improve generalization:

\textbf{1. Relational focus}:
Concept-enhanced policies prioritize transferable relational patterns (e.g., ``object inside container'') over surface features.
Our cross-modal correlation learning (Sec.~\ref{subsec:multi-modal}) enables this capability by identifying patterns that remain invariant across modalities.
This relational emphasis explains the stronger performance on scenarios that alter visual appearance while preserving task structure.
For instance, while \textbf{Novel Placement} tests spatial variation alone,
the other protocols introduce substantial visual perturbations (different colors, objects, or occlusions) that shift the appearance distribution.
The consistent performance gains across these visually diverse scenarios (Tab.~\ref{tab:real-exp-result}) suggest that the learned concepts successfully capture the underlying relational invariant
—placing cups into containers
—rather than memorizing superficial visual patterns.

\textbf{2. Hierarchical awareness}:
Concept-enhanced policies exhibit more systematic failure recovery than baselines, suggesting better tracking of sub-goal completion.
Baseline failures frequently exhibit premature task abandonment: the robot moves toward containers without having grasped objects,
or hovers near placement locations without executing placement.
In contrast, when concept-enhanced policies fail initial grasp attempts, they consistently retry grasping (typically 2-3 attempts) before proceeding, demonstrating recognition of incomplete sub-goals.
Although these recoveries ultimately fail due to time limits or object displacement, they reveal structured task progression rather than blind action execution.

\begin{wrapfigure}{r}{0.55\textwidth}
    \vspace{-4mm}
    \captionof{table}{\textbf{Real-world generalization success rates (\%)} for ACT policies with and without manipulation concepts (MC). Test conditions: Placements (novel layouts), Color (new pairings), Objects (unseen items), Obstacles (external barriers), Barrier (internal dividers), and Multi-grasp (two cups simultaneously).}
    \centering
    \begin{tabular}{ccccccc}
            \toprule
             & \rotatebox{60}{Place} & \rotatebox{60}{Color} & \rotatebox{60}{Obj.} & \rotatebox{60}{Obst.} & \rotatebox{60}{Barr.} & \rotatebox{60}{Multi} \\
            \midrule
            w/o MC & 53.3 & 46.7 & 40.0 & 20.0 & 0.0 & 0.0 \\
            w/ MC & \textbf{73.3} & \textbf{60.0} & \textbf{53.3} & \textbf{33.3} & \textbf{20.0} & \textbf{13.3} \\
            \bottomrule
    \end{tabular}
    \label{tab:real-exp-result}
    \vspace{-6mm}
\end{wrapfigure}

These mechanisms may enable manipulation concepts to promote policy generalization by encoding fundamental spatial and functional relationships that remain consistent across environmental variations.
Details are provided in Sec.~\ref{subsec:real-robot-exp-details}.

\vspace{-1mm}
\paragraph{Multi-Horizon Goal Prediction Visualization}
To visualize the temporal information encoded in our manipulation concepts, we examine outputs from our Multi-Horizon Goal Predictor (MHGP, $\mathcal{F}$ in Eq.~\ref{eq:loss_predictable goal}) using the BridgeDataV2 dataset \citep{walke2023bridgedata}. Fig.~\ref{fig:pred_and_recon_example} (Sec.~\ref{subsec:vis-pred-and-recon-details}) shows predicted goal states when conditioned on the current observation, manipulation concept, and various coherence thresholds ($\epsilon$).

The predictions capture essential task structures -- such as anticipated arm trajectories and object interactions -- rather than attempting pixel-perfect reconstructions. This abstraction of scene-specific details in favor of functional relationships is crucial for cross-environment generalization. Importantly, as $\epsilon$ increases, the predictions correspond to states progressively further into the future, with smaller values showing immediate next steps and larger values revealing final goal states. This demonstrates that our learned concepts encode meaningful temporal structures at multiple time horizons, enabling policies to simultaneously reason about immediate actions and longer-term objectives. Details are provided in Sec.~\ref{subsec:vis-pred-and-recon-details}.

\section{Discussion}
\label{sec:discussion}

We demonstrate that 
self-supervised discovery of hierarchical manipulation concepts significantly enhances robot policy performance across original tasks, novel compositions, and entirely new environments. 
Three key strengths emerge:
(1) our representations naturally resemble semantically meaningful manipulation primitives without requiring explicit labels, as evidenced by diagonal clustering in similarity matrices; 
(2) the concepts bridge low-level actions and high-level goals through hierarchical organization, enabling reasoning at multiple temporal scales;
and
(3) concept-enhanced policies focus on transferable relational patterns rather than superficial features, explaining their robust generalization to scenarios with substantial distribution shifts.
These findings highlight the potential of learning manipulation concepts from unlabeled multi-modal demonstrations for creating more adaptable and interpretable robotic systems. 
Limitations are discussed in Sec.~\ref{subsec:future works}.

\newpage

\begin{ack}
This work 
is supported by 
the Early Career Scheme 
of the Research Grants Council (RGC) grant \# 27207224,
the HKU-100 Award, 
a donation from the Musketeers Foundation, 
in part by the 
JC STEM Lab of Robotics for Soft Materials 
funded by The Hong Kong Jockey Club Charities Trust,
and DAMO Academy through the Alibaba Innovative Research Program.
\end{ack}

\bibliographystyle{plain}
\bibliography{reference}

\begin{thebibliography}{10}

\bibitem{axmacher2006memory}
Nikolai Axmacher, Florian Mormann, Guillen Fern{\'a}ndez, Christian~E Elger, and Juergen Fell.
\newblock Memory formation by neuronal synchronization.
\newblock {\em Brain research reviews}, 52(1):170--182, 2006.

\bibitem{belghazi2018mutual}
Mohamed~Ishmael Belghazi, Aristide Baratin, Sai Rajeshwar, Sherjil Ozair, Yoshua Bengio, Aaron Courville, and Devon Hjelm.
\newblock Mutual information neural estimation.
\newblock In {\em International conference on machine learning}, pages 531--540. PMLR, 2018.

\bibitem{belkhale2024rt}
Suneel Belkhale, Tianli Ding, Ted Xiao, Pierre Sermanet, Quon Vuong, Jonathan Tompson, Yevgen Chebotar, Debidatta Dwibedi, and Dorsa Sadigh.
\newblock Rt-h: Action hierarchies using language.
\newblock {\em arXiv preprint arXiv:2403.01823}, 2024.

\bibitem{black2024pi_0}
Kevin Black, Noah Brown, Danny Driess, Adnan Esmail, Michael Equi, Chelsea Finn, Niccolo Fusai, Lachy Groom, Karol Hausman, Brian Ichter, et~al.
\newblock $\pi_0$: A vision-language-action flow model for general robot control.
\newblock {\em arXiv preprint arXiv:2410.24164}, 2024.

\bibitem{blackzero}
Kevin Black, Mitsuhiko Nakamoto, Pranav Atreya, Homer~Rich Walke, Chelsea Finn, Aviral Kumar, and Sergey Levine.
\newblock Zero-shot robotic manipulation with pre-trained image-editing diffusion models.
\newblock In {\em The Twelfth International Conference on Learning Representations}, 2024.

\bibitem{bonatti2023pact}
Rogerio Bonatti, Sai Vemprala, Shuang Ma, Felipe Frujeri, Shuhang Chen, and Ashish Kapoor.
\newblock Pact: Perception-action causal transformer for autoregressive robotics pre-training.
\newblock In {\em 2023 IEEE/RSJ International Conference on Intelligent Robots and Systems (IROS)}, pages 3621--3627. IEEE, 2023.

\bibitem{bruce2024genie}
Jake Bruce, Michael~D Dennis, Ashley Edwards, Jack Parker-Holder, Yuge Shi, Edward Hughes, Matthew Lai, Aditi Mavalankar, Richie Steigerwald, Chris Apps, et~al.
\newblock Genie: Generative interactive environments.
\newblock In {\em Forty-first International Conference on Machine Learning}, 2024.

\bibitem{bu2024towards}
Qingwen Bu, Hongyang Li, Li~Chen, Jisong Cai, Jia Zeng, Heming Cui, Maoqing Yao, and Yu~Qiao.
\newblock Towards synergistic, generalized, and efficient dual-system for robotic manipulation.
\newblock {\em arXiv preprint arXiv:2410.08001}, 2024.

\bibitem{chi2023diffusion}
Cheng Chi, Zhenjia Xu, Siyuan Feng, Eric Cousineau, Yilun Du, Benjamin Burchfiel, Russ Tedrake, and Shuran Song.
\newblock Diffusion policy: Visuomotor policy learning via action diffusion.
\newblock {\em arXiv preprint arXiv:2303.04137}, 2023.

\bibitem{oxe}
Embodiment Collaboration, Abby O'Neill, Abdul Rehman, Abhinav Gupta, Abhiram Maddukuri, Abhishek Gupta, Abhishek Padalkar, Abraham Lee, Acorn Pooley, Agrim Gupta, Ajay Mandlekar, Ajinkya Jain, Albert Tung, Alex Bewley, Alex Herzog, Alex Irpan, Alexander Khazatsky, Anant Rai, Anchit Gupta, Andrew Wang, Andrey Kolobov, Anikait Singh, Animesh Garg, Aniruddha Kembhavi, Annie Xie, Anthony Brohan, Antonin Raffin, Archit Sharma, Arefeh Yavary, Arhan Jain, Ashwin Balakrishna, Ayzaan Wahid, Ben Burgess-Limerick, Beomjoon Kim, Bernhard Schölkopf, Blake Wulfe, Brian Ichter, Cewu Lu, Charles Xu, Charlotte Le, Chelsea Finn, Chen Wang, Chenfeng Xu, Cheng Chi, Chenguang Huang, Christine Chan, Christopher Agia, Chuer Pan, Chuyuan Fu, Coline Devin, Danfei Xu, Daniel Morton, Danny Driess, Daphne Chen, Deepak Pathak, Dhruv Shah, Dieter Büchler, Dinesh Jayaraman, Dmitry Kalashnikov, Dorsa Sadigh, Edward Johns, Ethan Foster, Fangchen Liu, Federico Ceola, Fei Xia, Feiyu Zhao, Felipe~Vieira Frujeri, Freek Stulp, Gaoyue Zhou,
  Gaurav~S. Sukhatme, Gautam Salhotra, Ge~Yan, Gilbert Feng, Giulio Schiavi, Glen Berseth, Gregory Kahn, Guangwen Yang, Guanzhi Wang, Hao Su, Hao-Shu Fang, Haochen Shi, Henghui Bao, Heni~Ben Amor, Henrik~I Christensen, Hiroki Furuta, Homanga Bharadhwaj, Homer Walke, Hongjie Fang, Huy Ha, Igor Mordatch, Ilija Radosavovic, Isabel Leal, Jacky Liang, Jad Abou-Chakra, Jaehyung Kim, Jaimyn Drake, Jan Peters, Jan Schneider, Jasmine Hsu, Jay Vakil, Jeannette Bohg, Jeffrey Bingham, Jeffrey Wu, Jensen Gao, Jiaheng Hu, Jiajun Wu, Jialin Wu, Jiankai Sun, Jianlan Luo, Jiayuan Gu, Jie Tan, Jihoon Oh, Jimmy Wu, Jingpei Lu, Jingyun Yang, Jitendra Malik, João Silvério, Joey Hejna, Jonathan Booher, Jonathan Tompson, Jonathan Yang, Jordi Salvador, Joseph~J. Lim, Junhyek Han, Kaiyuan Wang, Kanishka Rao, Karl Pertsch, Karol Hausman, Keegan Go, Keerthana Gopalakrishnan, Ken Goldberg, Kendra Byrne, Kenneth Oslund, Kento Kawaharazuka, Kevin Black, Kevin Lin, Kevin Zhang, Kiana Ehsani, Kiran Lekkala, Kirsty Ellis, Krishan Rana,
  Krishnan Srinivasan, Kuan Fang, Kunal~Pratap Singh, Kuo-Hao Zeng, Kyle Hatch, Kyle Hsu, Laurent Itti, Lawrence~Yunliang Chen, Lerrel Pinto, Li~Fei-Fei, Liam Tan, Linxi~"Jim" Fan, Lionel Ott, Lisa Lee, Luca Weihs, Magnum Chen, Marion Lepert, Marius Memmel, Masayoshi Tomizuka, Masha Itkina, Mateo~Guaman Castro, Max Spero, Maximilian Du, Michael Ahn, Michael~C. Yip, Mingtong Zhang, Mingyu Ding, Minho Heo, Mohan~Kumar Srirama, Mohit Sharma, Moo~Jin Kim, Naoaki Kanazawa, Nicklas Hansen, Nicolas Heess, Nikhil~J Joshi, Niko Suenderhauf, Ning Liu, Norman~Di Palo, Nur Muhammad~Mahi Shafiullah, Oier Mees, Oliver Kroemer, Osbert Bastani, Pannag~R Sanketi, Patrick~"Tree" Miller, Patrick Yin, Paul Wohlhart, Peng Xu, Peter~David Fagan, Peter Mitrano, Pierre Sermanet, Pieter Abbeel, Priya Sundaresan, Qiuyu Chen, Quan Vuong, Rafael Rafailov, Ran Tian, Ria Doshi, Roberto Mart'in-Mart'in, Rohan Baijal, Rosario Scalise, Rose Hendrix, Roy Lin, Runjia Qian, Ruohan Zhang, Russell Mendonca, Rutav Shah, Ryan Hoque, Ryan Julian,
  Samuel Bustamante, Sean Kirmani, Sergey Levine, Shan Lin, Sherry Moore, Shikhar Bahl, Shivin Dass, Shubham Sonawani, Shubham Tulsiani, Shuran Song, Sichun Xu, Siddhant Haldar, Siddharth Karamcheti, Simeon Adebola, Simon Guist, Soroush Nasiriany, Stefan Schaal, Stefan Welker, Stephen Tian, Subramanian Ramamoorthy, Sudeep Dasari, Suneel Belkhale, Sungjae Park, Suraj Nair, Suvir Mirchandani, Takayuki Osa, Tanmay Gupta, Tatsuya Harada, Tatsuya Matsushima, Ted Xiao, Thomas Kollar, Tianhe Yu, Tianli Ding, Todor Davchev, Tony~Z. Zhao, Travis Armstrong, Trevor Darrell, Trinity Chung, Vidhi Jain, Vikash Kumar, Vincent Vanhoucke, Wei Zhan, Wenxuan Zhou, Wolfram Burgard, Xi~Chen, Xiangyu Chen, Xiaolong Wang, Xinghao Zhu, Xinyang Geng, Xiyuan Liu, Xu~Liangwei, Xuanlin Li, Yansong Pang, Yao Lu, Yecheng~Jason Ma, Yejin Kim, Yevgen Chebotar, Yifan Zhou, Yifeng Zhu, Yilin Wu, Ying Xu, Yixuan Wang, Yonatan Bisk, Yongqiang Dou, Yoonyoung Cho, Youngwoon Lee, Yuchen Cui, Yue Cao, Yueh-Hua Wu, Yujin Tang, Yuke Zhu, Yunchu
  Zhang, Yunfan Jiang, Yunshuang Li, Yunzhu Li, Yusuke Iwasawa, Yutaka Matsuo, Zehan Ma, Zhuo Xu, Zichen~Jeff Cui, Zichen Zhang, Zipeng Fu, and Zipeng Lin.
\newblock Open x-embodiment: Robotic learning datasets and rt-x models, 2024.

\bibitem{dasari2023unbiased}
Sudeep Dasari, Mohan~Kumar Srirama, Unnat Jain, and Abhinav Gupta.
\newblock An unbiased look at datasets for visuo-motor pre-training.
\newblock In {\em Conference on Robot Learning}, pages 1183--1198. PMLR, 2023.

\bibitem{ding2023task}
Yan Ding, Xiaohan Zhang, Chris Paxton, and Shiqi Zhang.
\newblock Task and motion planning with large language models for object rearrangement.
\newblock In {\em 2023 IEEE/RSJ International Conference on Intelligent Robots and Systems (IROS)}, pages 2086--2092. IEEE, 2023.

\bibitem{eisner2022flowbot3d}
Ben Eisner, Harry Zhang, and David Held.
\newblock Flowbot3d: Learning 3d articulation flow to manipulate articulated objects.
\newblock {\em arXiv preprint arXiv:2205.04382}, 2022.

\bibitem{fangsam2act}
Haoquan Fang, Markus Grotz, Wilbert Pumacay, Yi~Ru Wang, Dieter Fox, Ranjay Krishna, and Jiafei Duan.
\newblock Sam2act: Integrating visual foundation model with a memory architecture for robotic manipulation.
\newblock In {\em 7th Robot Learning Workshop: Towards Robots with Human-Level Abilities}.

\bibitem{fell2011role}
Juergen Fell and Nikolai Axmacher.
\newblock The role of phase synchronization in memory processes.
\newblock {\em Nature reviews neuroscience}, 12(2):105--118, 2011.

\bibitem{fu2024mobile}
Zipeng Fu, Tony~Z Zhao, and Chelsea Finn.
\newblock Mobile aloha: Learning bimanual mobile manipulation with low-cost whole-body teleoperation.
\newblock {\em arXiv preprint arXiv:2401.02117}, 2024.

\bibitem{grafton2007evidence}
Scott~T Grafton and Antonia F de~C Hamilton.
\newblock Evidence for a distributed hierarchy of action representation in the brain.
\newblock {\em Human movement science}, 26(4):590--616, 2007.

\bibitem{guo2024prediction}
Yanjiang Guo, Yucheng Hu, Jianke Zhang, Yen-Jen Wang, Xiaoyu Chen, Chaochao Lu, and Jianyu Chen.
\newblock Prediction with action: Visual policy learning via joint denoising process.
\newblock In {\em The Thirty-eighth Annual Conference on Neural Information Processing Systems}, 2024.

\bibitem{intelligence2025pi}
Physical Intelligence, Kevin Black, Noah Brown, James Darpinian, Karan Dhabalia, Danny Driess, Adnan Esmail, Michael Equi, Chelsea Finn, Niccolo Fusai, et~al.
\newblock $\pi_{0.5}$: a vision-language-action model with open-world generalization.
\newblock {\em arXiv preprint arXiv:2504.16054}, 2025.

\bibitem{jia2024chainofthought}
Zhiwei Jia, Vineet Thumuluri, Fangchen Liu, Linghao Chen, Zhiao Huang, and Hao Su.
\newblock Chain-of-thought predictive control.
\newblock In {\em Forty-first International Conference on Machine Learning}, 2024.

\bibitem{jia2024chain}
Zhiwei Jia, Vineet Thumuluri, Fangchen Liu, Linghao Chen, Zhiao Huang, and Hao Su.
\newblock Chain-of-thought predictive control.
\newblock In {\em International Conference on Machine Learning}, pages 21768--21790. PMLR, 2024.

\bibitem{karamcheti2023language}
Siddharth Karamcheti, Suraj Nair, Annie~S Chen, Thomas Kollar, Chelsea Finn, Dorsa Sadigh, and Percy Liang.
\newblock Language-driven representation learning for robotics.
\newblock {\em arXiv preprint arXiv:2302.12766}, 2023.

\bibitem{kim2025fine}
Moo~Jin Kim, Chelsea Finn, and Percy Liang.
\newblock Fine-tuning vision-language-action models: Optimizing speed and success.
\newblock {\em arXiv preprint arXiv:2502.19645}, 2025.

\bibitem{kim2024openvla}
Moo~Jin Kim, Karl Pertsch, Siddharth Karamcheti, Ted Xiao, Ashwin Balakrishna, Suraj Nair, Rafael Rafailov, Ethan Foster, Grace Lam, Pannag Sanketi, et~al.
\newblock Openvla: An open-source vision-language-action model.
\newblock {\em arXiv preprint arXiv:2406.09246}, 2024.

\bibitem{kipf2019compile}
Thomas Kipf, Yujia Li, Hanjun Dai, Vinicius Zambaldi, Alvaro Sanchez-Gonzalez, Edward Grefenstette, Pushmeet Kohli, and Peter Battaglia.
\newblock Compile: Compositional imitation learning and execution.
\newblock In {\em International Conference on Machine Learning}, pages 3418--3428. PMLR, 2019.

\bibitem{lee2024behavior}
Seungjae Lee, Yibin Wang, Haritheja Etukuru, H~Jin Kim, Nur Muhammad~Mahi Shafiullah, and Lerrel Pinto.
\newblock Behavior generation with latent actions.
\newblock {\em arXiv preprint arXiv:2403.03181}, 2024.

\bibitem{li2024decisionnce}
Jianxiong Li, Jinliang Zheng, Yinan Zheng, Liyuan Mao, Xiao Hu, Sijie Cheng, Haoyi Niu, Jihao Liu, Yu~Liu, Jingjing Liu, et~al.
\newblock Decisionnce: Embodied multimodal representations via implicit preference learning.
\newblock In {\em International Conference on Machine Learning}, pages 29461--29488. PMLR, 2024.

\bibitem{li2023manipllm}
Xiaoqi Li, Mingxu Zhang, Yiran Geng, Haoran Geng, Yuxing Long, Yan Shen, Renrui Zhang, Jiaming Liu, and Hao Dong.
\newblock Manipllm: Embodied multimodal large language model for object-centric robotic manipulation.
\newblock {\em arXiv preprint arXiv:2312.16217}, 2023.

\bibitem{liang2024skilldiffuser}
Zhixuan Liang, Yao Mu, Hengbo Ma, Masayoshi Tomizuka, Mingyu Ding, and Ping Luo.
\newblock Skilldiffuser: Interpretable hierarchical planning via skill abstractions in diffusion-based task execution.
\newblock In {\em Proceedings of the IEEE/CVF Conference on Computer Vision and Pattern Recognition}, pages 16467--16476, 2024.

\bibitem{LIBERO}
Bo~Liu, Yifeng Zhu, Chongkai Gao, Yihao Feng, Qiang Liu, Yuke Zhu, and Peter Stone.
\newblock Libero: Benchmarking knowledge transfer for lifelong robot learning.
\newblock In A.~Oh, T.~Naumann, A.~Globerson, K.~Saenko, M.~Hardt, and S.~Levine, editors, {\em Advances in Neural Information Processing Systems}, volume~36, pages 44776--44791. Curran Associates, Inc., 2023.

\bibitem{liuinfocon}
Ruizhe Liu, Qian Luo, and Yanchao Yang.
\newblock Infocon: Concept discovery with generative and discriminative informativeness.
\newblock In {\em The Twelfth International Conference on Learning Representations}, 2024.

\bibitem{liu2025one}
Yuyao Liu, Jiayuan Mao, Joshua~B Tenenbaum, Tom{\'a}s Lozano-P{\'e}rez, and Leslie~Pack Kaelbling.
\newblock One-shot manipulation strategy learning by making contact analogies.
\newblock In {\em 2025 IEEE International Conference on Robotics and Automation (ICRA)}, pages 15387--15393. IEEE, 2025.

\bibitem{lynch2020learning}
Corey Lynch, Mohi Khansari, Ted Xiao, Vikash Kumar, Jonathan Tompson, Sergey Levine, and Pierre Sermanet.
\newblock Learning latent plans from play.
\newblock In {\em Conference on robot learning}, pages 1113--1132. PMLR, 2020.

\bibitem{ma2023liv}
Yecheng~Jason Ma, Vikash Kumar, Amy Zhang, Osbert Bastani, and Dinesh Jayaraman.
\newblock Liv: Language-image representations and rewards for robotic control.
\newblock In {\em International Conference on Machine Learning}, pages 23301--23320. PMLR, 2023.

\bibitem{ma2022vip}
Yecheng~Jason Ma, Shagun Sodhani, Dinesh Jayaraman, Osbert Bastani, Vikash Kumar, and Amy Zhang.
\newblock Vip: Towards universal visual reward and representation via value-implicit pre-training.
\newblock {\em arXiv preprint arXiv:2210.00030}, 2022.

\bibitem{majumdar2023we}
Arjun Majumdar, Karmesh Yadav, Sergio Arnaud, Jason Ma, Claire Chen, Sneha Silwal, Aryan Jain, Vincent-Pierre Berges, Tingfan Wu, Jay Vakil, et~al.
\newblock Where are we in the search for an artificial visual cortex for embodied intelligence?
\newblock {\em Advances in Neural Information Processing Systems}, 36:655--677, 2023.

\bibitem{mao2023learning}
Jiayuan Mao, Tom{\'a}s Lozano-P{\'e}rez, Joshua~B Tenenbaum, and Leslie~Pack Kaelbling.
\newblock Learning reusable manipulation strategies.
\newblock In {\em Conference on Robot Learning}, pages 1467--1483. PMLR, 2023.

\bibitem{melloni2007synchronization}
Lucia Melloni, Carlos Molina, Marcela Pena, David Torres, Wolf Singer, and Eugenio Rodriguez.
\newblock Synchronization of neural activity across cortical areas correlates with conscious perception.
\newblock {\em Journal of neuroscience}, 27(11):2858--2865, 2007.

\bibitem{mete2024quest}
Atharva Mete, Haotian Xue, Albert Wilcox, Yongxin Chen, and Animesh Garg.
\newblock Quest: Self-supervised skill abstractions for learning continuous control.
\newblock {\em arXiv preprint arXiv:2407.15840}, 2024.

\bibitem{mo2019partnet}
Kaichun Mo, Shilin Zhu, Angel~X Chang, Li~Yi, Subarna Tripathi, Leonidas~J Guibas, and Hao Su.
\newblock Partnet: A large-scale benchmark for fine-grained and hierarchical part-level 3d object understanding.
\newblock In {\em Proceedings of the IEEE/CVF conference on computer vision and pattern recognition}, pages 909--918, 2019.

\bibitem{murray2014hierarchy}
John~D Murray, Alberto Bernacchia, David~J Freedman, Ranulfo Romo, Jonathan~D Wallis, Xinying Cai, Camillo Padoa-Schioppa, Tatiana Pasternak, Hyojung Seo, Daeyeol Lee, et~al.
\newblock A hierarchy of intrinsic timescales across primate cortex.
\newblock {\em Nature neuroscience}, 17(12):1661--1663, 2014.

\bibitem{nair2022r3m}
Suraj Nair, Aravind Rajeswaran, Vikash Kumar, Chelsea Finn, and Abhinav Gupta.
\newblock R3m: A universal visual representation for robot manipulation.
\newblock {\em arXiv preprint arXiv:2203.12601}, 2022.

\bibitem{oquab2023dinov2}
Maxime Oquab, Timoth{\'e}e Darcet, Th{\'e}o Moutakanni, Huy Vo, Marc Szafraniec, Vasil Khalidov, Pierre Fernandez, Daniel Haziza, Francisco Massa, Alaaeldin El-Nouby, et~al.
\newblock Dinov2: Learning robust visual features without supervision.
\newblock {\em arXiv preprint arXiv:2304.07193}, 2023.

\bibitem{pertsch2020keyframing}
Karl Pertsch, Oleh Rybkin, Jingyun Yang, Shenghao Zhou, Konstantinos Derpanis, Kostas Daniilidis, Joseph Lim, and Andrew Jaegle.
\newblock Keyframing the future: Keyframe discovery for visual prediction and planning.
\newblock In {\em Learning for Dynamics and Control}, pages 969--979. PMLR, 2020.

\bibitem{pertsch2025fast}
Karl Pertsch, Kyle Stachowicz, Brian Ichter, Danny Driess, Suraj Nair, Quan Vuong, Oier Mees, Chelsea Finn, and Sergey Levine.
\newblock Fast: Efficient action tokenization for vision-language-action models.
\newblock {\em arXiv preprint arXiv:2501.09747}, 2025.

\bibitem{clip}
Alec Radford, Jong~Wook Kim, Chris Hallacy, Aditya Ramesh, Gabriel Goh, Sandhini Agarwal, Girish Sastry, Amanda Askell, Pamela Mishkin, Jack Clark, et~al.
\newblock Learning transferable visual models from natural language supervision.
\newblock In {\em International conference on machine learning}, pages 8748--8763. PMLR, 2021.

\bibitem{radosavovic2023robot}
Ilija Radosavovic, Baifeng Shi, Letian Fu, Ken Goldberg, Trevor Darrell, and Jitendra Malik.
\newblock Robot learning with sensorimotor pre-training.
\newblock In {\em Conference on Robot Learning}, pages 683--693. PMLR, 2023.

\bibitem{rholanguage}
Seungeun Rho, Laura Smith, Tianyu Li, Sergey Levine, Xue~Bin Peng, and Sehoon Ha.
\newblock Language guided skill discovery.
\newblock In {\em The Thirteenth International Conference on Learning Representations}.

\bibitem{rombach2021highresolution}
Robin Rombach, Andreas Blattmann, Dominik Lorenz, Patrick Esser, and Björn Ommer.
\newblock High-resolution image synthesis with latent diffusion models, 2021.

\bibitem{schmidtlearning}
Dominik Schmidt and Minqi Jiang.
\newblock Learning to act without actions.
\newblock In {\em The Twelfth International Conference on Learning Representations}, 2024.

\bibitem{seo2023masked}
Younggyo Seo, Danijar Hafner, Hao Liu, Fangchen Liu, Stephen James, Kimin Lee, and Pieter Abbeel.
\newblock Masked world models for visual control.
\newblock In {\em Conference on Robot Learning}, pages 1332--1344. PMLR, 2023.

\bibitem{shafiullah2022behavior}
Nur~Muhammad Shafiullah, Zichen Cui, Ariuntuya~Arty Altanzaya, and Lerrel Pinto.
\newblock Behavior transformers: Cloning $ k $ modes with one stone.
\newblock {\em Advances in neural information processing systems}, 35:22955--22968, 2022.

\bibitem{shah2023mutex}
Rutav Shah, Roberto Mart{\'\i}n-Mart{\'\i}n, and Yuke Zhu.
\newblock Mutex: Learning unified policies from multimodal task specifications.
\newblock {\em arXiv preprint arXiv:2309.14320}, 2023.

\bibitem{Sharma2020Dynamics-Aware}
Archit Sharma, Shixiang Gu, Sergey Levine, Vikash Kumar, and Karol Hausman.
\newblock Dynamics-aware unsupervised discovery of skills.
\newblock In {\em International Conference on Learning Representations}, 2020.

\bibitem{sharma2023multiresolution}
Shashank Sharma, Vinay Namboodiri, and Janina Hoffmann.
\newblock Multi-resolution skill discovery for hierarchical reinforcement learning.
\newblock In {\em NeurIPS 2023 Workshop on Goal-Conditioned Reinforcement Learning}, 2023.

\bibitem{shi2023waypoint}
Lucy~Xiaoyang Shi, Archit Sharma, Tony~Z Zhao, and Chelsea Finn.
\newblock Waypoint-based imitation learning for robotic manipulation.
\newblock In {\em Conference on Robot Learning}, pages 2195--2209. PMLR, 2023.

\bibitem{singer2011consciousness}
Wolf Singer.
\newblock Consciousness and neuronal synchronization.
\newblock {\em The neurology of consciousness}, pages 43--52, 2011.

\bibitem{tian2025predictive}
Yang Tian, Sizhe Yang, Jia Zeng, Ping Wang, Dahua Lin, Hao Dong, and Jiangmiao Pang.
\newblock Predictive inverse dynamics models are scalable learners for robotic manipulation.
\newblock In {\em The Thirteenth International Conference on Learning Representations}, 2025.

\bibitem{van2017neural}
Aaron Van Den~Oord, Oriol Vinyals, et~al.
\newblock Neural discrete representation learning.
\newblock {\em Advances in neural information processing systems}, 30, 2017.

\bibitem{walke2023bridgedata}
Homer~Rich Walke, Kevin Black, Tony~Z Zhao, Quan Vuong, Chongyi Zheng, Philippe Hansen-Estruch, Andre~Wang He, Vivek Myers, Moo~Jin Kim, Max Du, et~al.
\newblock Bridgedata v2: A dataset for robot learning at scale.
\newblock In {\em Conference on Robot Learning}, pages 1723--1736. PMLR, 2023.

\bibitem{wan2024lotus}
Weikang Wan, Yifeng Zhu, Rutav Shah, and Yuke Zhu.
\newblock Lotus: Continual imitation learning for robot manipulation through unsupervised skill discovery.
\newblock In {\em 2024 IEEE International Conference on Robotics and Automation (ICRA)}, pages 537--544. IEEE, 2024.

\bibitem{wangscaling}
Lirui Wang, Xinlei Chen, Jialiang Zhao, and Kaiming He.
\newblock Scaling proprioceptive-visual learning with heterogeneous pre-trained transformers.
\newblock In {\em The Thirty-eighth Annual Conference on Neural Information Processing Systems}, 2024.

\bibitem{womelsdorf2007role}
Thilo Womelsdorf and Pascal Fries.
\newblock The role of neuronal synchronization in selective attention.
\newblock {\em Current opinion in neurobiology}, 17(2):154--160, 2007.

\bibitem{wu2023unleashing}
Hongtao Wu, Ya~Jing, Chilam Cheang, Guangzeng Chen, Jiafeng Xu, Xinghang Li, Minghuan Liu, Hang Li, and Tao Kong.
\newblock Unleashing large-scale video generative pre-training for visual robot manipulation.
\newblock {\em arXiv preprint arXiv:2312.13139}, 2023.

\bibitem{xu2023xskill}
Mengda Xu, Zhenjia Xu, Cheng Chi, Manuela Veloso, and Shuran Song.
\newblock Xskill: Cross embodiment skill discovery.
\newblock In {\em Conference on Robot Learning}, pages 3536--3555. PMLR, 2023.

\bibitem{yang2023polybot}
Jonathan Yang, Dorsa Sadigh, and Chelsea Finn.
\newblock Polybot: Training one policy across robots while embracing variability.
\newblock {\em arXiv preprint arXiv:2307.03719}, 2023.

\bibitem{yang2022chain}
Mengjiao~Sherry Yang, Dale Schuurmans, Pieter Abbeel, and Ofir Nachum.
\newblock Chain of thought imitation with procedure cloning.
\newblock {\em Advances in Neural Information Processing Systems}, 35:36366--36381, 2022.

\bibitem{ye2024latent}
Seonghyeon Ye, Joel Jang, Byeongguk Jeon, Sejune Joo, Jianwei Yang, Baolin Peng, Ajay Mandlekar, Reuben Tan, Yu-Wei Chao, Bill~Yuchen Lin, et~al.
\newblock Latent action pretraining from videos.
\newblock {\em arXiv preprint arXiv:2410.11758}, 2024.

\bibitem{zeng2024learning}
Jia Zeng, Qingwen Bu, Bangjun Wang, Wenke Xia, Li~Chen, Hao Dong, Haoming Song, Dong Wang, Di~Hu, Ping Luo, et~al.
\newblock Learning manipulation by predicting interaction.
\newblock {\em arXiv preprint arXiv:2406.00439}, 2024.

\bibitem{zhao2025cot}
Qingqing Zhao, Yao Lu, Moo~Jin Kim, Zipeng Fu, Zhuoyang Zhang, Yecheng Wu, Zhaoshuo Li, Qianli Ma, Song Han, Chelsea Finn, et~al.
\newblock Cot-vla: Visual chain-of-thought reasoning for vision-language-action models.
\newblock {\em arXiv preprint arXiv:2503.22020}, 2025.

\bibitem{zhaolearning}
Tony~Z. Zhao, Vikash Kumar, Sergey Levine, and Chelsea Finn.
\newblock Learning fine-grained bimanual manipulation with low-cost hardware.
\newblock In {\em ICML Workshop on New Frontiers in Learning, Control, and Dynamical Systems}, 2023.

\bibitem{zhen3d}
Haoyu Zhen, Xiaowen Qiu, Peihao Chen, Jincheng Yang, Xin Yan, Yilun Du, Yining Hong, and Chuang Gan.
\newblock 3d-vla: A 3d vision-language-action generative world model.
\newblock In {\em Forty-first International Conference on Machine Learning}, 2024.

\bibitem{zhouautocgp}
Pei Zhou, Ruizhe Liu, Qian Luo, Fan Wang, Yibing Song, and Yanchao Yang.
\newblock Autocgp: Closed-loop concept-guided policies from unlabeled demonstrations.
\newblock In {\em The Thirteenth International Conference on Learning Representations}.

\bibitem{zhu2022bottom}
Yifeng Zhu, Peter Stone, and Yuke Zhu.
\newblock Bottom-up skill discovery from unsegmented demonstrations for long-horizon robot manipulation.
\newblock {\em IEEE Robotics and Automation Letters}, 7(2):4126--4133, 2022.

\bibitem{robosuite2020}
Yuke Zhu, Josiah Wong, Ajay Mandlekar, Roberto Mart\'{i}n-Mart\'{i}n, Abhishek Joshi, Soroush Nasiriany, Yifeng Zhu, and Kevin Lin.
\newblock robosuite: A modular simulation framework and benchmark for robot learning.
\newblock In {\em arXiv preprint arXiv:2009.12293}, 2020.

\end{thebibliography}








\appendix

\section{Implementation details}\label{sec:Implementation details}
\subsection{Manipulation concept discovery (Ours)}\label{subsec:id_MCD}
This section details the neural network architectures and training procedures employed in our manipulation concepts discovery framework (Sec.~\ref{sec:method}) as implemented on the LIBERO benchmark.

\paragraph{Manipulation Concept Encoder (Sec.~\ref{subsec:mcd})}
The manipulation concepts encoder \(\mathcal{E}\) (Eq.~\ref{eq:concept_encoder}) first encodes the multi-modal observations at each time step of the input demonstration into an encoded vector. It then utilizes a self-attention transformer to process the sequence of encoded vectors into a sequence of manipulation concepts.  
For the observation encoding process, our experiments on LIBERO incorporate two vision observations: \textbf{agent-view vision} and \textbf{eye-in-hand vision}. The original images are tensors of shape \(128 \times 128 \times 3\). To enhance processing efficiency, we preprocess the images for each time step using the VAE encoder from stable diffusion \citep{rombach2021highresolution}, compressing each image into a tensor of shape \(16 \times 16 \times 4\), which is then flattened into a 1024-dimensional vector.  
In addition to the two vision observations, we include a 9-dimensional robot state at each time step of each demonstration as the proprioceptive state observation. For these three observations at each time step, we employ three distinct 2-layer MLPs to process each observation into a feature vector of the hidden size (256) used by the subsequent transformer. The encoded features from these observations are then summed to form a 256-dimensional representation that encapsulates the sensing information from the three modalities.
\begin{equation}
\begin{aligned}
& h_{\text{av}} = \mathrm{MLP}_\text{av}(
I_\text{av-compress})\quad
h_{\text{ev}} = \mathrm{MLP}_\text{ev}(
I_\text{ev-compress})\quad
h_{\text{prop}} = \mathrm{MLP}_\text{prop}(
s_\text{prop}) \\
& h = h_{\text{av}} + h_{\text{ev}} + h_{\text{prop}}
\end{aligned}
\label{eq:encode_obs}
\end{equation}
Here, \(I_\text{av-compress}\) represents the 1024-dimensional compressed \textbf{agent-view vision}, and \(I_\text{ev-compress}\) represents the 1024-dimensional compressed \textbf{eye-in-hand vision}. \(s_\text{prop}\) denotes the 9-dimensional proprioceptive state observation. The output of the hidden layers from the three MLPs is 1024 dimensions.
The \(h\) in Eq.~\ref{eq:encode_obs} represents the encoded observation feature at each time step of a given demonstration \(\tau_i\): \((h_i^1, h_i^2, \cdots, h_i^{T_i})\).  
The next module in \(\mathcal{E}\) is a 12-layer self-attention (MHA in Eq.~\ref{eq:encode_attn}) transformer,
enabling each time step to aggregate information from every other time step in the input sequence.
In our implementation, we do not input the entire demonstration; instead, the transformer processes a fixed input sequence length of \(T_\text{context} = 60\).
A learnable temporal embedding,
represented as a tensor of shape \(60 \times 256\),
is added to the input sequence to enhance temporal representation.
The hidden feature dimension at each time step is 256, and each self-attention layer consists of 8 heads.
Moreover, since spherical distance is utilized in Sec.~\ref{subsec:multi-hierarchy}, 
the output manipulation concepts are normalized to have a unit length with respect to the 2-norm:
\begin{equation}
\begin{aligned}
(z_i^t,z_i^{t+1},\cdots,z_i^{t+T_\text{context}-1}) \leftarrow \mathrm{Norm}_2\left(\left[\mathrm{MHA}\right]_{\times12}\left(h_i^t,h_i^{t+1},\cdots,h_i^{t+T_\text{context}-1}\right)\right)
\end{aligned}
\label{eq:encode_attn}
\end{equation}
The output manipulation concept sequence in Eq.~\ref{eq:concept_encoder} represents the predicted manipulation concepts at time-steps \(t, t+1, \cdots, (t+T_\text{context}-1)\) of the demonstration \(\tau_i\).  
During training, demonstrations with lengths shorter than \(T_\text{context}\) are padded to \(T_\text{context}\) by repeating the observations from the last time-step at the end of each demonstration.  
During inference,
when \(\mathcal{E}\) is used to label the demonstrations in the original dataset,
the manipulation concepts at each time step are designed to incorporate information from as many future time steps as possible.
This approach aims to better capture motion pattern dynamics,
aligning with prior works that generate the dynamics at the current time step based on information derived from the dynamics spanning the current to future time steps \citep{ye2024latent}.
Specifically:
\begin{itemize}[left=0em,itemsep=0pt, parsep=0pt]
    \item For each time-step \(t \leq T_i - T_\text{context}\),
    the corresponding manipulation concepts are derived when
    the input to Eq.~\ref{eq:encode_attn}
    starts from this time-step and spans a length of \(T_\text{context}\):
    \(\left(h_i^t, h_i^{t+1}, \cdots, h_i^{t+T_\text{context}-1}\right)\).
    \item For each time step \(t > T_i - T_\text{context}\),
    the corresponding manipulation concepts are derived when the input to Eq.~\ref{eq:encode_attn} begins at time step \(h_i^{T_i-T_\text{context}+1}\) and spans a length of \(T_\text{context}\),
    ensuring that the final time step corresponds to the end of the demonstration:
    \(\left(h_i^{T_i-T_\text{context}+1}, h_i^{T_i-T_\text{context}+2}, \cdots, h_i^{T_i}\right)\).
    \item If the original demonstration length is smaller than \(T_\text{context}\), the manipulation concepts correspond to the input appended with repeated observations as described earlier. 
\end{itemize}
However, we do not firmly believe this is the optimal approach for labeling manipulation concepts.
Further exploration of inference-time strategy design is left for future work,
as it is not a core focus of the manipulation concept discovery methodology presented.

\paragraph{Learning Multi-Modal Features and Correlations (Sec.~\ref{subsec:multi-modal})}
The Cross-Modal Correlation Network \(\mathcal{C}\) (Eq.~\ref{eq:loss_mi_and_moc}) shares a similar structure with \(\mathcal{E}\) (Eq.~\ref{eq:concept_encoder}).
First, it includes four 2-layer MLP encoders, analogous to the three encoders in Eq.~\ref{eq:encode_obs}, with an additional encoder for processing the manipulation concepts. Each of these four MLPs outputs a hidden feature of dimension 1024, which is then reduced to a 256-dimensional encoded feature. These encoded features are summed to represent the combined information from the three observations and the manipulation concepts.
Second, it incorporates a 4-layer self-attention transformer to process the sequence of features (with the same fixed length \(T_\text{context} = 60\)) produced by the four MLPs. Following this, three 3-layer MLP decoders map the transformer's output to the reconstructed observations at each time step. Unlike in Eq.~\ref{eq:encode_attn}, the transformer's output does not require normalization. Each decoder MLP has hidden layers with a dimension of 1024.
As described in Eq.~\ref{eq:loss_mi_and_moc}, for the three observations—\textbf{agent-view camera vision}, \textbf{eye-in-hand camera vision}, and \textbf{proprioceptive state observation}—we randomly mask these modalities, ensuring that at least one modality is masked during each iteration. The \(2^3 - 1 = 7\) possible masking scenarios follow a uniform distribution, with each scenario appearing with a probability of \(\frac{1}{7}\). For the sampled masks, all observations of the corresponding masked modalities in the input sequence are replaced with zero tensors.
The loss is applied separately to the reconstruction of the three different observations. Specifically, L2 loss is applied to the two vision observations, while L1 loss is applied to the proprioceptive state observations.
The loss weight \(\lambda_\text{mm}\) in Eq.~\ref{eq:joint_mcd_loss} is set to 1.0.

\paragraph{Learning Multi-Hierarchical Sub-goals (Sec.~\ref{subsec:multi-hierarchy})}
The Multi-Horizon Future Predictor \(\mathcal{F}\) (Eq.~\ref{eq:loss_predictable goal}) shares a similar structure with \(\mathcal{C}\) (Eq.~\ref{eq:loss_mi_and_moc}). The key differences are:
\begin{itemize}[left=0em,itemsep=0pt, parsep=0pt]
    \item \(\mathcal{F}\) does not require a masking strategy.
    \item The transformer in \(\mathcal{F}\) is a 4-layer causal self-attention transformer. Causal attention is used because, in Eq.~\ref{eq:loss_predictable goal}, the prediction is made from each current time step to certain future time steps. Therefore, for each time-step input in \(\mathcal{F}\), access to information from subsequent time steps is restricted.

    \item To incorporate the granularity parameter \(\epsilon \in [0,1]\), we discretize the continuous range into 1000 uniform bins \(\{0.000, 0.001, \ldots, 0.999\}\) and learn a corresponding VQ-VAE codebook \citep{van2017neural} with 1000 entries, each represented as a 256-dimensional embedding vector. In each transformer block, the feed-forward layer receives the concatenation of the attention output and the embedding corresponding to the sampled \(\epsilon\) value.
    
    \item The output predictions correspond to the observations at the time steps determined by the rules described in Sec.~\ref{subsec:multi-hierarchy} (Eqs.~\ref{eq:t_structure} and~\ref{eq:t_terminal}).
    Still, the loss is applied separately to the reconstruction of the three types of observations. Specifically, L2 loss is used for the two vision observations, while L1 loss is applied to the proprioceptive state observations.
The loss weight \(\lambda_\text{mh}\) in Eq.~\ref{eq:joint_mcd_loss} is set to 1.0.
\end{itemize}

\paragraph{Training Details}
We train the manipulation concept discovery process for 200,000 iterations with a batch size of 512.
Each item in the batch is a segment of demonstration with a fixed length of \(T_\text{context} = 60\).
The training process uses the AdamW optimizer with a weight decay of 0.001 and momentum parameters \(\beta_1 = 0.9\) and \(\beta_2 = 0.95\).
The base learning rate is set to 0.001.
Initially, the model is trained with a 100-iteration warmup phase,
during which the learning rate increases linearly from 0.0001 to 0.001.
After the warmup, the model is trained for the remaining iterations using a cosine decay schedule,
gradually reducing the learning rate back to 0.0001.
This training setup is compatible with GPUs such as the GeForce RTX 3090 or 4090.
However, we leverage the A800 GPU for improved efficiency, completing the training process in 1.5 days.

\subsection{Enhancing Imitation Learning}\label{subsec:id_EIL}

\begin{figure}[ht]
\centering
\includegraphics[width=1.0\linewidth]{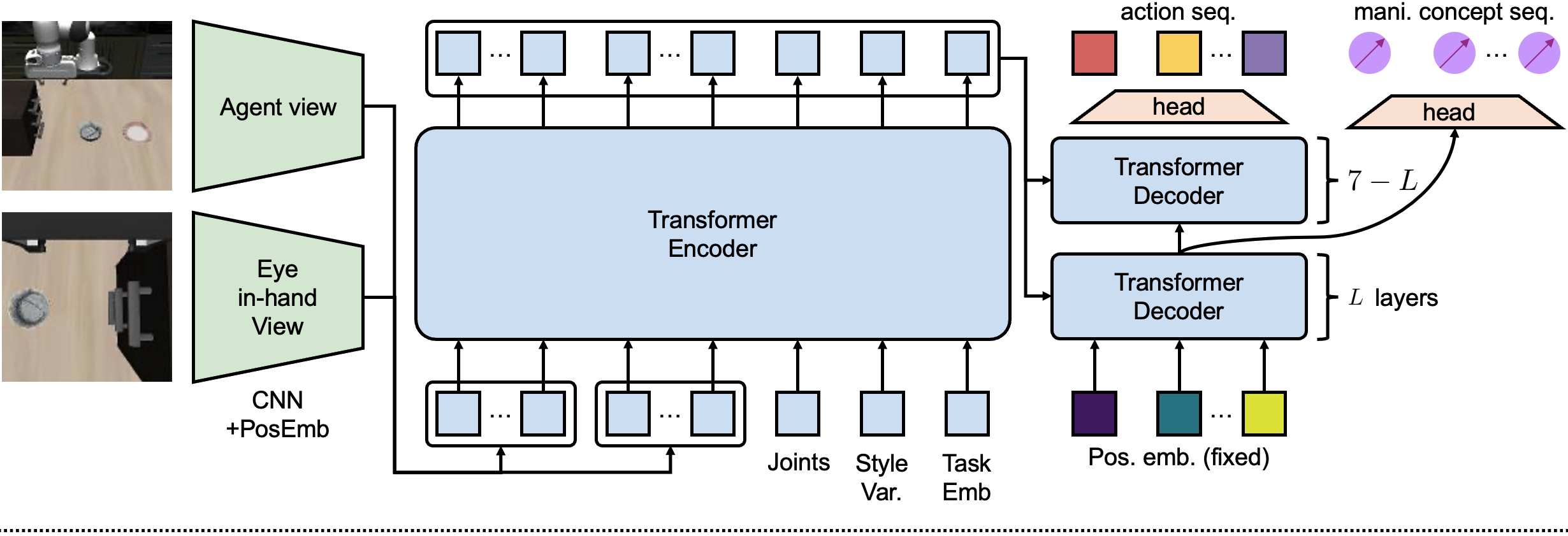}
\includegraphics[width=0.7\linewidth]{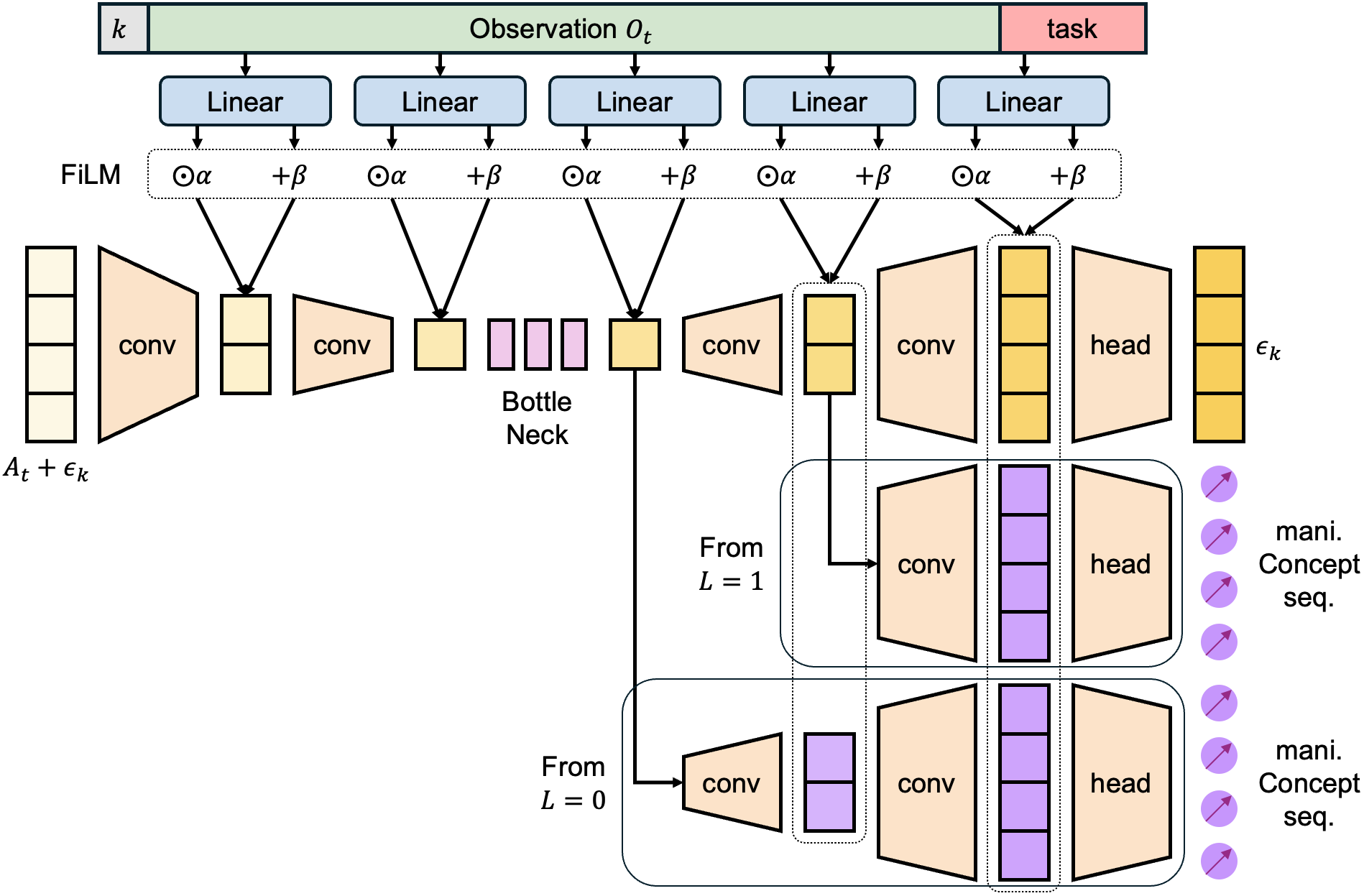}
\caption{Upper: Enhanced ACT (decoder part); lower: Enhanced Diffusion Policy}
\label{fig:ACT_and_DP}
\end{figure}

This section introduces the neural network architectures and training details used in the Enhancing Imitation Learning process (Sec.~\ref{subsec:EIL}).
It focuses on modifying the original neural network policy to enable the prediction of manipulation concepts,
thereby enhancing performance.
Moreover, the implementation of the base policy follow \citep{mete2024quest}.

\paragraph{ACT}
The pipeline (we focus on the CVAE decoder as it is the only modified component)
is shown in the upper part of Fig.~\ref{fig:ACT_and_DP}.
Following \citep{mete2024quest}, the transformer encoder in ACT's CVAE decoder is modified to incorporate task embeddings provided by CLIP.
The transformer decoder in ACT's CVAE decoder is adapted to predict manipulation concepts.  
Specifically, the output of the \(L\)-th layer in the transformer decoder is processed by an additional decoding head, which is nearly identical to the one used for outputting action chunks, with the only modification being the output dimension. This decoding head outputs manipulation concept chunks corresponding to the same time steps as the action chunks, with its parameters adjusted to match the dimensionality of the manipulation concepts at each time step (256). Other training and testing settings follow \citep{mete2024quest}.  
Moreover, the transformer decoder in ACT's CVAE decoder, as implemented by \citep{mete2024quest}, consists of 7 layers. During our experiments, we tested various combinations of \(L\)-th layers to determine the optimal layer for processing by the manipulation concept decoding head. Our results indicate that \(L=2\) provides slightly better performance than other configurations.
We present the ablation study on \(L\) and the weight \(\lambda_\text{mc}\) in Eq.~\ref{eq:policy_align_mc} for ACT on LIBERO-90 tasks, as shown in Tab.~\ref{tab:layer-weight-ACT}
\footnote{We provide sample rollouts (\texttt{supplementary/rollout\_summary}) and .gif logs of test-time ACT and DP performance (\texttt{supplementary/rollout\_video\_samples\_gif}) in the supplementary materials.}.
However, we believe this raises an interesting and challenging direction for future work: systematically investigating the rationale and insights behind the selection of \(L\), even beyond the context of our setting.

\begin{table}[ht]
\vspace{-3mm}
\caption{Ablation study on the intermediate layer outputs (\(L\)) used as inputs to the manipulation concept decoder and the loss weight \(\lambda_\text{mc}\) in Eq.~\ref{eq:policy_align_mc} for Enhancing Imitation Learning in ACT, evaluated on LIBERO-90 tasks.}
\centering
\begin{tabular}{cccccccc}
\toprule
ACT & \(\lambda_\text{mc}=1.0\) & \(\lambda_\text{mc}=0.1\) & \(\lambda_\text{mc}=0.01\) & \(\lambda_\text{mc}=0.001\) \\
\midrule
\(L=2\) & \textbf{74.8±0.8} & 70.6±0.8 & 69.0±0.1 & 68.7±0.5 \\
\midrule
\(L=3\) & 70.0±0.4 & 69.9±0.2 & 68.8±1.0 & 68.7±0.6 \\
\midrule
\(L=4\) & 72.6±0.5 & 69.9±0.3 & 69.6±0.2 & 67.3±0.5 \\
\bottomrule
\end{tabular}
\label{tab:layer-weight-ACT}
\vspace{-2mm}
\end{table}

\paragraph{Diffusion Policy}
The pipeline is illustrated in the lower part of Fig.~\ref{fig:ACT_and_DP}. Following \citep{mete2024quest}, the convolution-based Diffusion Policy is modified to concatenate the noise level (\(k\) and corresponding \(\epsilon_k\)) embedding, observation, and task embedding as the conditional input to the diffusion model network, using the FiLM strategy.  
We further introduce an additional manipulation concept decoding up-sampling module, nearly identical to the one used for outputting action chunks, with the only modification being the output dimension, to decode intermediate outputs from the corresponding up-sampling layer of the diffusion model. This decoding head can be configured to process intermediate outputs to predict manipulation concept chunks corresponding to the time steps of the predicted (noise of) action chunk outputs. The figure illustrates the cases for \(L=0\) and \(L=1\).
During our experiments, we tested various combinations of \(L\)-th layers to identify the optimal layer for processing by the manipulation concept decoding head. Our results suggest that \(L=1\) achieves better performance than other configurations.
We present the ablation study on \(L\) and the weight \(\lambda_\text{mc}\) in Eq.~\ref{eq:policy_align_mc} for Diffusion Policy on LIBERO-90 tasks, as shown in Tab.~\ref{tab:layer-weight-DP}.
Similar to ACT, we believe this topic needs further systematic study to uncover deeper insights.
Other training and testing settings follow \citep{mete2024quest}.

\begin{table}[ht]
\caption{Ablation study on the intermediate layer outputs (\(L\)) used as inputs to the manipulation concept decoder and the loss weight \(\lambda_\text{mc}\) in Eq.~\ref{eq:policy_align_mc} for Enhancing Imitation Learning in Diffusion Policy, evaluated on LIBERO-90 tasks.}
\centering
\begin{tabular}{cccccccc}
\toprule
DP & \(\lambda_\text{mc}=1.0\) & \(\lambda_\text{mc}=0.1\) & \(\lambda_\text{mc}=0.01\) & \(\lambda_\text{mc}=0.001\) \\
\midrule
\(L=0\) & 83.5±0.8 & 78.9±0.4 & 78.7±0.3 & 75.6±0.6 \\
\midrule
\(L=1\) & 80.0±0.4 & \textbf{89.6±0.6} & 82.0±0.2 & 79.9±0.1 \\
\bottomrule
\end{tabular}
\label{tab:layer-weight-DP}
\end{table}

Our future work includes a deeper study of modification strategies for various policies to adapt to the Enhancing Imitation Learning framework, following the methodology outlined in Sec.~\ref{subsec:EIL}.

\subsection{Manipulation concept discovery (Baselines)}\label{subsec:id_baselines}
\begin{itemize}[left=0em,itemsep=0pt, parsep=0pt]
    \item \textbf{InfoCon.}
    Based on the design of InfoCon \citep{liuinfocon},
    All the size of hidden features output by transformers and concept features is $256$.
    The state encoder (also process video clips consisting of concatenated,
    compressed vision observations and proprioceptive states, as outlined in Sec.~\ref{subsec:id_MCD}) uses a 12-layer transformer.
    The state reconstructor uses a 4-layer transformer.
    The goal-based policy uses a 4-layer transformer.
    The predictor for the generative goal uses a 4-layer transformer.
    For hyper-network used for discriminative goals, we use 2 hidden layers in the goal function. The number of concepts is fixed, the maximum number of $30$ manipulation concepts for all the tasks.
    We employ the AdamW optimizer, coupled with a warm-up cosine annealing scheduler same as Sec.~\ref{subsec:id_MCD}.
    The weight decay is always $1.0 \times 10^{-3}$.
    We use a batch size of $512$ during training.
    We train our model for 200,000 iterations with a base learning rate of \(1.0 \times 10^{-3}\) on a single A800 GPU within 1.5 days.
    \item \textbf{XSkill.}
    Following the design of XSkill~\cite{xu2023xskill},
    we implement its skill discovery framework on LIBERO-90,
    focusing exclusively on the ``robot'' embodiment and the Skill Discovery component from the XSkill pipeline.
    To ensure comparable model capacity and support multi-modality,
    our implementation employs a 12-layer Transformer as the temporal skill encoder.
    This encoder processes video clips consisting of concatenated,
    compressed vision observations and proprioceptive states,
    as outlined in Sec.~\ref{subsec:id_MCD},
    along with a trainable token to predict skill representations,
    which are subsequently used for skill prototype prediction.
    To augment the concatenated video clips containing multi-modality information,
    Gaussian noise with \(\sigma=1.0 \times 10^{-3}\) is applied.
    This unified augmentation approach accommodates the nature of proprioceptive states,
    as standard image augmentation techniques are not directly suitable for robotic proprioception.
    The training process employs a batch size of 512 and a learning rate of \(1.0 \times 10^{-3}\) for 200,000 iterations on a single A800 GPU within 1.5 days.
    \item \textbf{DecisionNCE} We fine-tune the DecisionNCE-T model (\url{https://github.com/2toinf/DecisionNCE}) on our dataset, as it outperforms DecisionNCE-P in our analysis of the experimental results in \citep{li2024decisionnce}. We use two types of language annotations: (1) the original task descriptions (Decision-task), and (2) detailed subtask labels derived by decomposing each task into meaningful subprocesses (Decision-motion). To construct the latter, we manually segment each demonstration based on changes in the robot’s proprioceptive state (e.g., movement direction, gripper open/close status). Segments corresponding to the same task are then assigned unified subtask labels across demonstrations, with remaining inconsistencies resolved through manual adjustment.
    \item \textbf{RPT.}
    We modify the original RPT design \citep{radosavovic2023robot} to adapt it for our task of discovering manipulation concept latents in the LIBERO-90 setting.
    We employ a 16-layer self-attention transformer to process inputs consisting of 60 consecutive, interleaved agent-view and eye-in-hand vision frames.
    Vision inputs are compressed using a stable diffusion VAE encoder, similar to the method in Sec.~\ref{subsec:id_MCD}.
    The total sequence length processed by the transformer is \(60 \times 3 = 180\).
    Each modality is mapped to a 256-dimensional embedding vector using an MLP, as defined in Eq.~\ref{eq:encode_obs}.
    The transformer’s output is then decoded to reconstruct the original inputs using a 3-layer MLP with 1024-dimensional hidden layers.
    We follow the masking strategy outlined in \citep{radosavovic2023robot} to perform temporal MAE training for the transformer.
    To label manipulation concept latents using the trained transformer,
    we extract the intermediate output of the 12th layer when the input consists of the full observation without masking.
    Notice that we select the output at the proprioceptive state input positions of the transformer to represent the manipulation concept latent at each time-step.
    The labeling process follows the procedure introduced in Sec.~\ref{subsec:id_MCD}.
    For training, we use a batch size of 512 and a learning rate of \(1.0 \times 10^{-3}\), running for 200,000 iterations on a single A800 GPU, which completes within 3 days.
    \item \textbf{All.}
    This is an ablation version of our manipulation concept discovery method,
    focusing on the design for capturing multi-modal correlations (Sec.~\ref{subsec:multi-modal}).
    Specifically, this baseline replaces the loss in Eq.~\ref{eq:loss_mi_and_moc} with a loss that does not use partial masking but instead always masks all modalities:  
    \(\mathcal{L}_{\text{all}}\left(t,\tau_i\right)
    =\left\| \mathcal{C}\left(\emptyset \Big| z^t_i;\Theta_c\right) - \textbf{o}^t_i \right\|\). Based on our reasoning and the experiment results show in Tab.~\ref{tab:mi}, we think this method may not be good at learning correlation between different modalities.
    Other settings follow Sec.~\ref{subsec:id_MCD}.
    \item \textbf{Next.}
    This is an ablation version of our manipulation concept discovery method,
    focusing on the design for representing multi-horizon subgoals (Sec.~\ref{subsec:multi-hierarchy}).
    Specifically, this baseline replaces the loss in Eq.~\ref{eq:loss_predictable goal} with a loss that always predicts the next adjacent time-step observation: 
    \(\mathcal{L}_{\text{next}}\left(t, \tau_i\right)
    = \left\| \mathcal{F}\left(\mathbf{o}^t_i, z^t_i; \Theta_f\right) - \mathbf{o}^{t+1}_i \right\|\).
    We observe that this setting is commonly used in recent works \citep{bruce2024genie,ye2024latent},
    which learn representations based on adjacent time-step observations or observations separated by a fixed time horizon.
    We suggest that learning based on a fixed time horizon is conceptually similar to adjacent time-step settings,
    as the fixed time horizon can be interpreted as a unified time step.
    Our method differs by considering the temporal correlation across multiple variable horizons,
    which is also addressed by baseline methods like RPT.
    Other settings follow Sec.~\ref{subsec:id_MCD}.
    \item \textbf{CLIP.} To ensure compatibility with other baselines, which have an output dimension of 256, we select the \texttt{ViT-B/32 CLIP} model from the original source (\url{https://github.com/openai/CLIP}). This model outputs a 512-dimensional feature vector, the closest to 256 among the accessible CLIP models from this codebase when given an image.
    \item \textbf{DINOv2.} To match the output dimension of 256 used by other baselines, we select the \texttt{dinov2-small} DINOv2 model from the source at \url{https://huggingface.co/facebook/dinov2-small}.
    This model produces a 384-dimensional feature vector when given an image.
\end{itemize}

Note that DecisionNCE, CLIP, and DINOv2 baselines use only vision (and language) information for concept discovery.
We preserve their original modality structure rather than adapting them to include proprioceptive states,
as this would deviate from their pretraining foundations.

\subsection{Mutual information estimation}\label{subsec:id_MI}
The estimation of mutual information is based on MINE \citep{belghazi2018mutual}, which uses batchwise samples drawn from a joint distribution and employs a neural network to estimate the mutual information. To extend this approach for estimating conditional mutual information (CMI), we reformulate CMI by decomposing it into mutual information terms, as shown below:

\begin{equation}
\begin{aligned}
\mathbb{I}(X:Y \mid Z) =
\mathbb{I}(X:Y)
+\mathbb{I}(XY:Z)
-\mathbb{I}(X:Z)
-\mathbb{I}(Y:Z),
\end{aligned}
\label{eq:decompose_cond_MI}
\end{equation}

where \(XY\) denotes the random variable sampled from the joint distribution of \(X\) and \(Y\) and is represented as the concatenation of their encoded vectors. The neural network in MINE has two layers, with the hidden layer size set to 1.5 times of the dimensions of the two random variables.

\section{Pseudocode}
Here we provide pseudocode for
(i) Deriving subprocess from manipulation concept latents (Alg.~\ref{alg:e_cluster}).
(ii) Manipulation concept disocovery training process of our method (Alg.~\ref{alg:mcd_pseudocode}).

\section{More Study on Learned Manipulation Concepts}\label{sec:more exp}

\subsection{Additional Experiments on Enhanced Imitation Learning}\label{subsec:more eil exp}

\paragraph{Sampling Strategies}
In this part,
we focus on methodology for deriving hierarchical structures from learned representations (Sec.~\ref{subsec:multi-hierarchy}).
While we adopt a threshold-based hierarchy derivation method (Eq.~\ref{eq:t_structure}) as a proof of concept,
we acknowledge that alternative derivation methodologies warrant further investigation (see Sec.~\ref{subsec:future works}).
For the threshold-based approach,
we employ uniform sampling of the threshold $\epsilon$ during training.
This choice ensures full coverage of all possible hierarchical structures, as we do not know a priori which threshold values might be suboptimal.
To validate this design choice, we conduct an ablation study comparing different sampling strategies for $\epsilon$ in Eq.~\ref{eq:loss_predictable goal}:

\begin{table}[ht]
    \caption{\textbf{Sampling Strategies Ablation}. We compare different sampling strategies for $\epsilon$ in Eq.~\ref{eq:loss_predictable goal}. Manipulation concepts are learned from LIBERO-90 and applied to policy learning on LIBERO-90. We report success rates (\%).}
    \centering
    \begin{tabular}{cccc}
        \toprule
        \textbf{Sampling Strategy} & \textbf{Description} & \textbf{ACT} & \textbf{DP} \\
        \midrule
        Uniform (Ours) & $\epsilon\sim\mathcal{U}\big(0,1\big)$ & 74.8±0.8 & 89.6±0.6  \\
        \midrule
        Sparse & $\epsilon\sim\{0.1,0.2,\cdots,1.0\}$ & 67.6±0.5 & 81.1±0.8 \\
        \midrule
        Biased & $\epsilon\sim\mathcal{U}\big(\frac{1}{3},\frac{2}{3}\big)$ & 65.6±0.7 & 78.7±0.4 \\
        \bottomrule
    \end{tabular}
    \label{tab:sampling strategy}
\end{table}

As shown in Tab.~\ref{tab:sampling strategy},
uniform sampling currently achieves the best performance across both policy architectures.
We hypothesize that while task-specific sampling strategies might excel on particular subsets,
uniform sampling provides robust performance across diverse tasks due to its comprehensive coverage of the threshold space.
Future work could explore adaptive sampling strategies tailored to specific task distributions.

\paragraph{Learning Methodology Contribution}
We conduct an ablation study to isolate the contributions of our two core learning methodologies:
Capturing Multi-Modal Correlations (Sec.~\ref{subsec:multi-modal})
and
Representing Multi-Horizon Sub-Goals (Sec.~\ref{subsec:multi-hierarchy}). Tab.~\ref{tab:methodology ablation} compares three configurations:
(1) \textit{Cross-modal only}: learning with only cross-modal alignment objectives in Eq.~\ref{eq:loss_mi_and_moc},
(2) \textit{Multi-horizon only}: learning with multi-horizon sub-goal prediction in Eq.~\ref{eq:loss_predictable goal} but without cross-modal alignment,
and
(3) \textit{Full method}: combining both cross-modal alignment and multi-horizon prediction.

\begin{table}[ht]
    \caption{\textbf{Methodology Contribution Ablation}.
    We evaluate the contribution of each learning component by training manipulation concepts on LIBERO-90 and applying them to policy learning on LIBERO-90.
    We report success rates (\%).}
    \centering
    \begin{tabular}{ccc}
        \toprule
        \textbf{Method} & \textbf{ACT} & \textbf{DP} \\
        \midrule
        Cross-modal only & 69.1±0.6 & 82.8±1.0  \\
        \midrule
        Multi-horizon only & 71.6±0.4 & 80.5±0.5  \\
        \midrule
        Ours (Full method) & 74.8±0.8 & 89.6±0.6 \\
        \bottomrule
    \end{tabular}
    \label{tab:methodology ablation}
\end{table}

The results in Tab.~\ref{tab:methodology ablation} reveal that both components make substantial and complementary contributions to performance.
We attribute this synergy to the distinct roles of each component:
cross-modal alignment grounds the understanding of correlations across different modalities, while multi-horizon prediction captures hierarchical temporal structure. Together, they enable the learning of manipulation concepts that are both correlationally coherent and temporally structured, leading to more robust policy learning.

\paragraph{Data Constraint Experiments}
We evaluate whether manipulation concepts can help mitigate the challenges of imitation learning under limited data.
Specifically, we vary the amount of data available for training both the manipulation concept encoder (Eq.~\ref{eq:concept_encoder}) and the enhanced imitation learning framework (Sec.~\ref{subsec:EIL}) to assess their impact on policy success rates.
We conduct experiments on LIBERO-90 tasks using the diffusion policy.
As shown in Tab.~\ref{tab:constraint-learning-data},
incorporating manipulation concepts consistently improves policy performance compared to settings without them,
even under restricted data conditions.
This demonstrates that learning and leveraging manipulation concepts can make imitation learning more data-efficient and effective.

\begin{table}[ht]
    \caption{\textbf{Performance under data constraints.}
    Success rates of diffusion policies with and without manipulation concept enhancement, evaluated on LIBERO-90 (\textbf{\textit{L90-90}}). In each setting, the number of demonstrations per task available for training both the manipulation concept encoder and the policy is limited as indicated.}
    \centering
    \begin{tabular}{cccc}
        \toprule
         & \textbf{50 demos/task} & \textbf{25 demos/task} & \textbf{10 demos/task} \\
        \midrule
        Ours & 89.6 ± 0.6 & 77.6 ± 0.5 & 61.2 ± 1.1 \\
        \midrule
        Plain & 75.1 ± 0.6 & 70.1 ± 0.3 & 59.1 ± 0.9 \\
        \bottomrule
    \end{tabular}
    \label{tab:constraint-learning-data}
\end{table}

\paragraph{Distance Metric}
We conduct an ablation study comparing spherical distance and cosine distance $\frac{1-\cos{(\cdot})}{2}$ for $\text{dist}(\cdot,\cdot)$ in Eq.~\ref{eq:t_structure}. Tab.~\ref{tab:distance metric} reports the performance when concepts are learned and applied to LIBERO-90 tasks.
Further investigation into distance-threshold-based subprocess derivation methods represents a promising direction for future work.

\begin{table}[ht]
    \caption{Ablation study on distance metrics for concept learning on LIBERO-90. Spherical distance consistently outperforms cosine distance across both baseline methods.}
    \centering
    \begin{tabular}{ccc}
    \toprule
     & \textbf{Cosine Distance} & \textbf{Spherical Distance (Ours)} \\
    \midrule
    ACT & 67.8±0.5 & 74.8±0.8 \\
    \midrule
    DP & 82.0±0.4 & 89.6±0.6 \\
    \bottomrule
    \end{tabular}
    \label{tab:distance metric}
\end{table}

\paragraph{Sub-process Derivation}
We conduct an ablation study comparing two approaches for constraining manipulation concept latents within each sub-process in Eq.~\ref{eq:t_structure}.
Our proposed method enforces proximity among all concept latents throughout the sub-process (``Sequential Constraint''),
while the baseline only constrains the distance between the initial and final concept latents (``Endpoint Constraint'').
We evaluate both approaches on LIBERO-90, where concept discovery and policy enhancement are performed.
Tab.~\ref{tab:subprocess derivation ablation} reports the task success rates when integrating the learned manipulation concepts with different policy architectures.

\begin{table}[ht]
\caption{Ablation study on sub-process derivation constraints. We compare enforcing proximity among all manipulation concept latents within each sub-process (Sequential Constraint) versus constraining only the initial and final latents (Endpoint Constraint). Results show average success rates (\%) with standard errors across LIBERO-90 tasks.}
\centering
\begin{tabular}{ccc}
\toprule
 & \textbf{Sequential Constraint} & \textbf{Endpoint Constraint} \\
\midrule
ACT & 74.8±0.8 & 68.4±0.8 \\
\midrule
DP & 89.6±0.6 & 79.8±0.5 \\
\bottomrule
\end{tabular}
\label{tab:subprocess derivation ablation}
\end{table}


\paragraph{Future Prediction Strategy}
Apart from the different sub-goal determination strategies we compared
(\textbf{Next} and \textbf{InfoCon} in Sec.~\ref{subsec:exp_setups}),
we evaluate two additional future prediction strategies.

\begin{itemize}[left=0em,itemsep=0pt, parsep=0pt]
    \item \textbf{Next-n.} Unlike our sub-process derivation strategy (Eq.~\ref{eq:t_structure}), this baseline encodes future observations at varying time horizons by randomly sampling a future timestep. Specifically: \(\mathcal{L}_{\text{next-n}}\left(t, \tau_i\right) = \mathbb{E}_{n\sim\text{U}\{1,2,\cdots,T_i-t\}}\left\| \mathcal{F}\left(\mathbf{o}^t_i,z^t_i,n; \Theta_f\right) - \mathbf{o}^{t+n}_i \right\|\).
    \item \textbf{Next-random.} This strategy builds upon \textbf{Next-n} but differs in how future targets are selected. We first randomly segment training demonstrations into sub-processes for concept discovery. Then, for a state at time-step $t$, the prediction target is randomly selected from among the end-states of subsequent sub-processes. For example, if a demonstration is segmented into 5 sub-processes and time-step $t$ is in the 2nd sub-process, the model will randomly predict one of the end-states from the 2nd, 3rd, 4th, or 5th sub-processes during concept discovery learning.
\end{itemize}

We evaluated diffusion policies enhanced by these strategies,
with results presented in Tab.~\ref{tab:more-future}.
The data demonstrates that our manipulation concepts yield better policy enhancement compared to the alternative strategies.
This highlights the importance of carefully designing which future observations to predict and validates the effectiveness of our self-supervised sub-goal derivation and learning method.
Specifically, the performance decrease observed with \textbf{Next-n} and \textbf{Next-random},
despite their consideration of multi-horizon futures,
likely stems from the fact that not all future states effectively represent sub-goal completion.
Intermediate movement states may be reached through multiple alternative trajectories that ultimately achieve the same sub-goal,
thus providing limited information about the underlying task structure.

\begin{table}[ht]
    \caption{\textbf{Comparison of Additional Future Prediction Strategies.}
    Success rates of diffusion policies enhanced with manipulation concepts discovered using our method versus two alternative future prediction strategies on the LIBERO-90 benchmark.}
    \centering
    \begin{tabular}{cccc}
        \toprule
        \textbf{\textit{L90-90}} & \textbf{Ours} & \textbf{Next-n} & \textbf{Next-random} \\
        \midrule
        DP & 89.6 ± 0.6 & 83.0±0.3 & 82.8 ± 0.4 \\
        \bottomrule
    \end{tabular}
    \label{tab:more-future}
\end{table}

\paragraph{Usage of Manipulation Concept Encoder}

We investigate two strategies for leveraging the manipulation concept encoder from Eq.~\ref{eq:concept_encoder} in downstream policy learning. The encoder serves as an intermediate module that extracts manipulation concept representations from demonstrations. We compare the following approaches:
(1) \textbf{Direct Conditioning}: The trained encoder directly processes current observations to generate manipulation concepts, which are then concatenated with observations as additional input features to the policy network.
(2) \textbf{Joint Prediction (Ours)}: The policy network is trained to jointly predict both future actions and future manipulation concepts from current observations, as described in Sec.~\ref{subsec:EIL}.
Tab.~\ref{tab:direct_versus_joint} presents the comparative results across two policy architectures.

\begin{table}[ht]
    \caption{\textbf{Comparison of Manipulation Concept Usage Strategies.}}
    \centering
    \begin{tabular}{ccc}
        \toprule
        \textbf{Policy} & \textbf{Direct Conditioning} & \textbf{Joint Prediction (Ours)} \\
        \midrule
        ACT & 71.1±0.4 & 74.8±0.8 \\
        \midrule
        DP & 79.3±0.9 & 89.6±0.6 \\
        \bottomrule
    \end{tabular}
    \label{tab:direct_versus_joint}
\end{table}

The performance gap stems from a temporal alignment mismatch between concept representations and action predictions.
In \textbf{Direct Conditioning},
the encoder extracts concepts from current or past observations, creating a temporal lag: the policy receives historical concept information when planning future actions.
In contrast, Joint Prediction enforces temporal coherence by training the policy to predict future manipulation concepts alongside future actions,
ensuring that the predicted concepts align temporally with the planned action sequence.

This temporal alignment is critical in multi-phase manipulation tasks.
For example, consider a pick-and-place scenario:
immediately after grasping an object, the current observation encodes grasping-related dynamics.
However, to execute the subsequent placement action,
the policy requires placement-relevant information.
Joint Prediction learns to anticipate these future task-phase concepts,
providing the policy with forward-looking contextual information.
Direct Conditioning, by contrast, conditions the policy on backward-looking grasping concepts that offer limited guidance for placement planning.

While our results demonstrate the advantages of temporal alignment through joint prediction, we acknowledge that direct conditioning on historical concepts may benefit tasks requiring explicit long-horizon memory or reactive behaviors based on past states \citep{fangsam2act}. Future work will explore hybrid architectures that combine both strategies.

\subsection{Alignment with Semantic Sub-Goals}\label{subsec:semantic study}
We evaluate whether the manipulation concept latents learned by our method resemble human-interpretable semantics.
Specifically, we assess whether latents assigned to time steps of demonstrations (Sec.~\ref{subsec:mcd}) exhibit higher pairwise similarity when those steps belong to sub-processes pursuing the same human-defined sub-goal.

To analyze the learned representations,
we first group manipulation concept latents according to human-annotated sub-goals.
For instance, in the task ``open the top drawer'',
latents from time steps where the robot reaches for the top drawer handle are categorized as ``reach the top drawer''.
Latents from other demonstrations and tasks involving identical processes (reaching the top drawer) are placed in the same category.
We then quantify the similarity between two categories by calculating the average cosine similarity between their respective latents, as defined in Eq.~\ref{eq:class_sim}.

\begin{figure}[t!]
\centering
\includegraphics[width=0.8\linewidth]{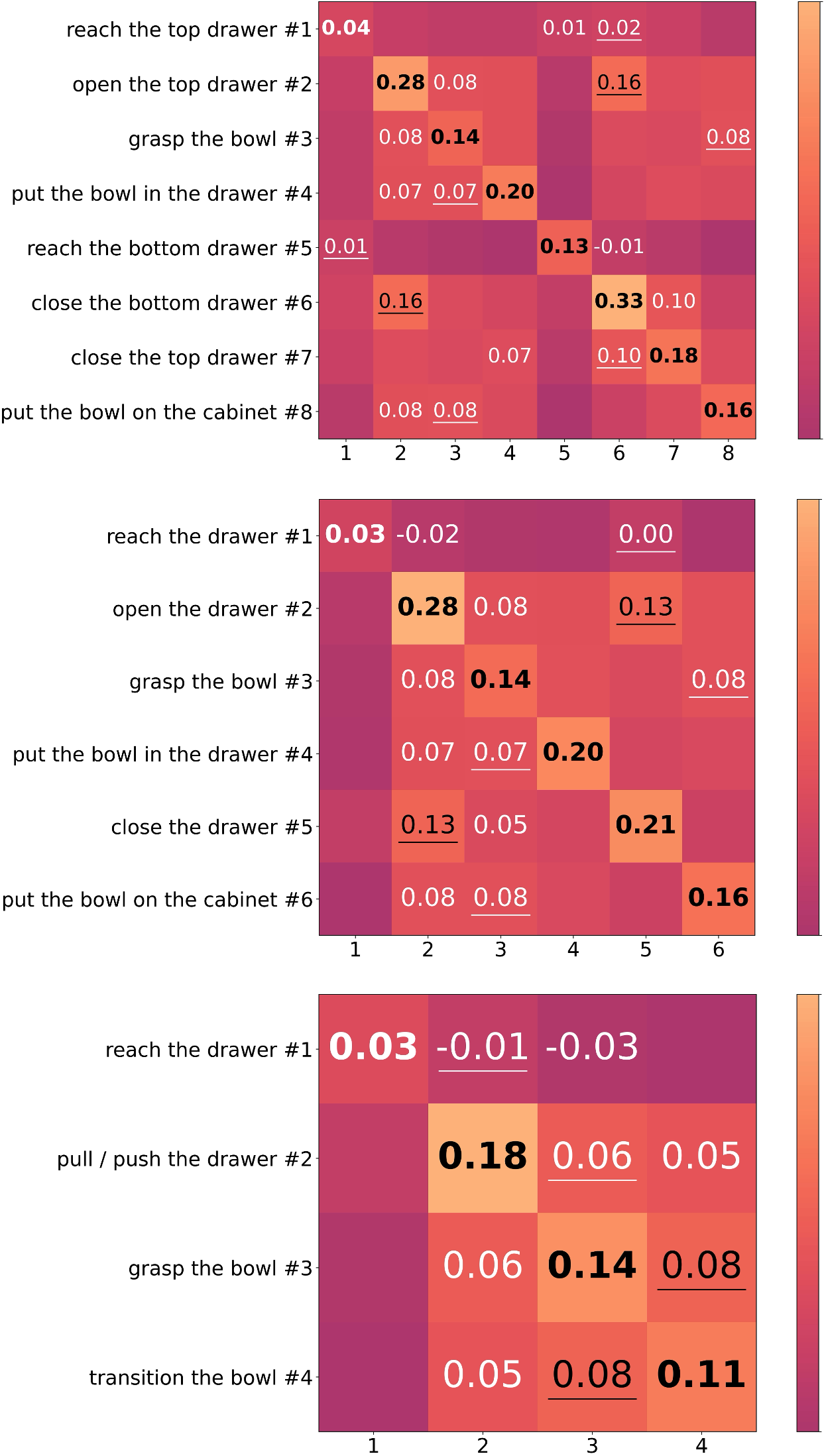}
\caption{
Average cosine similarity
between pairs of sub-goal categories (defined by human semantics)
computed using manipulation concept latents
learned by our method (Sec.\ref{sec:method}).
In each heatmap, the value at the \(i\)-th row and \(j\)-th column represents the average cosine similarity between latent vectors from the \(i\)-th and \(j\)-th categories. Three levels of labeling are provided across the heatmaps; please refer to Sec.~\ref{subsec:semantic study} for details.
}
\label{fig:semantic cluster}
\end{figure}

Fig.~\ref{fig:semantic cluster} shows results from analyzing demonstrations from three tasks:
\begin{itemize}[left=0em,itemsep=0pt, parsep=0pt]
\item Task \#1: Open the top drawer of the cabinet and put the bowl in it;
\item Task \#2: Close the bottom drawer of the cabinet and open the top drawer;
\item Task \#3: Close the top drawer of the cabinet and put the black bowl on top. 
\end{itemize}
We selected these tasks because they clearly demonstrate overlapping subgoals across different tasks
(e.g., Task \#1 and Task \#2 both include ``opening the top drawer'').
This enables testing whether the latents capture similar subgoal semantics across different tasks—an essential capability for cross-task learning efficiency (Sec.~\ref{sec:introduction}).
Manipulation concept latents are grouped based on human-defined sub-goals,
with similarities between category pairs visualized as heatmaps.
Three heatmaps are presented, each using a different granularity of sub-goal annotation:
\begin{itemize}[left=1em,itemsep=0pt, parsep=0pt] 
\item[1.]  Top-1st heatmap: Omits task-specific distinctions, merging similar manipulation processes across tasks into the same category  
\item[2.]  Top-2nd heatmap: Further merges similar manipulation processes, disregarding distinctions like ``top drawer'' versus ``bottom drawer''  
\item[3.]  Top-3rd heatmap: Consolidates manipulation processes further, treating actions like bowl transitions as the same concept regardless of context
\end{itemize}
In each heatmap, the entry at position $(i,j)$ represents the average similarity ($\times 10.0$) between categories $i$ and $j$. For readability, only the top three similarity values in each row are displayed.

We emphasize that testing semantic capture at different ``description granularity levels'' is important
because semantics naturally exist at multiple levels of abstraction,
from highly specific details to broadly generalizable patterns.
Finer-grained descriptions provide more precise details but limited generalization,
while coarser-grained descriptions capture more general features applicable across diverse scenarios.
For example,
the general instruction ``close the drawer'' applies broadly to subprocesses in both Task \#2 and Task \#3,
whereas the more specific ``close the top drawer'' incorporates spatial features that make it applicable in Task \#3 but not in Task \#2.
Through this multi-granularity analysis,
we evaluate whether our manipulation concept latents successfully capture
both fine-grained semantics needed for specific scenarios
and coarse-grained semantics that enable transfer across more scenarios.

What we observe is that the highest similarity values consistently appear along the diagonal in each heatmap in Fig.~\ref{fig:semantic cluster},
so concept latents from the same category show higher similarity compared with different categories.
This indicates that the learned latent clusters resemble clusters derived from human-interpretable sub-goal classifications,
suggesting that our model captures meaningful semantic structure in the manipulation processes.
Moreover, the patterns observed across the three heatmaps with different description granularities
reveal that the latents encode semantics at multiple levels of abstraction.
They capture both generalizable semantics applicable across tasks and scenes,
while simultaneously preserving fine-grained scene-specific details.

Furthermore,
Fig.~\ref{fig:tsne-task} provides a t-SNE visualization of manipulation concept latents from all 90 tasks in LIBERO-90.
For each task, latents ($z_i^t$) were extracted at every time step of demonstrations.
In the plot, latents are color-coded by their originating tasks.
We observe that clusters often contain latents from diverse tasks,
as indicated by the mixed colors in each cluster.
This further supports our finding that the learned latents generalize across tasks and capture shared semantic structures.\footnote{It should be noted that t-SNE performs extreme dimensionality reduction, so these clusters may not perfectly reflect similarity in the high-dimensional space. This visualization should therefore be considered as supplementary evidence.}

\subsection{Motion Study}\label{subsec:motion study}
\begin{figure}[!t]
    \centering
    \subfigure[Average cosine similarity
    between pairs of movement categories (defined by human semantics)
    computed using manipulation concept latents
    learned by our method (Sec.\ref{sec:method}).]{
        \includegraphics[width=0.38\textwidth]{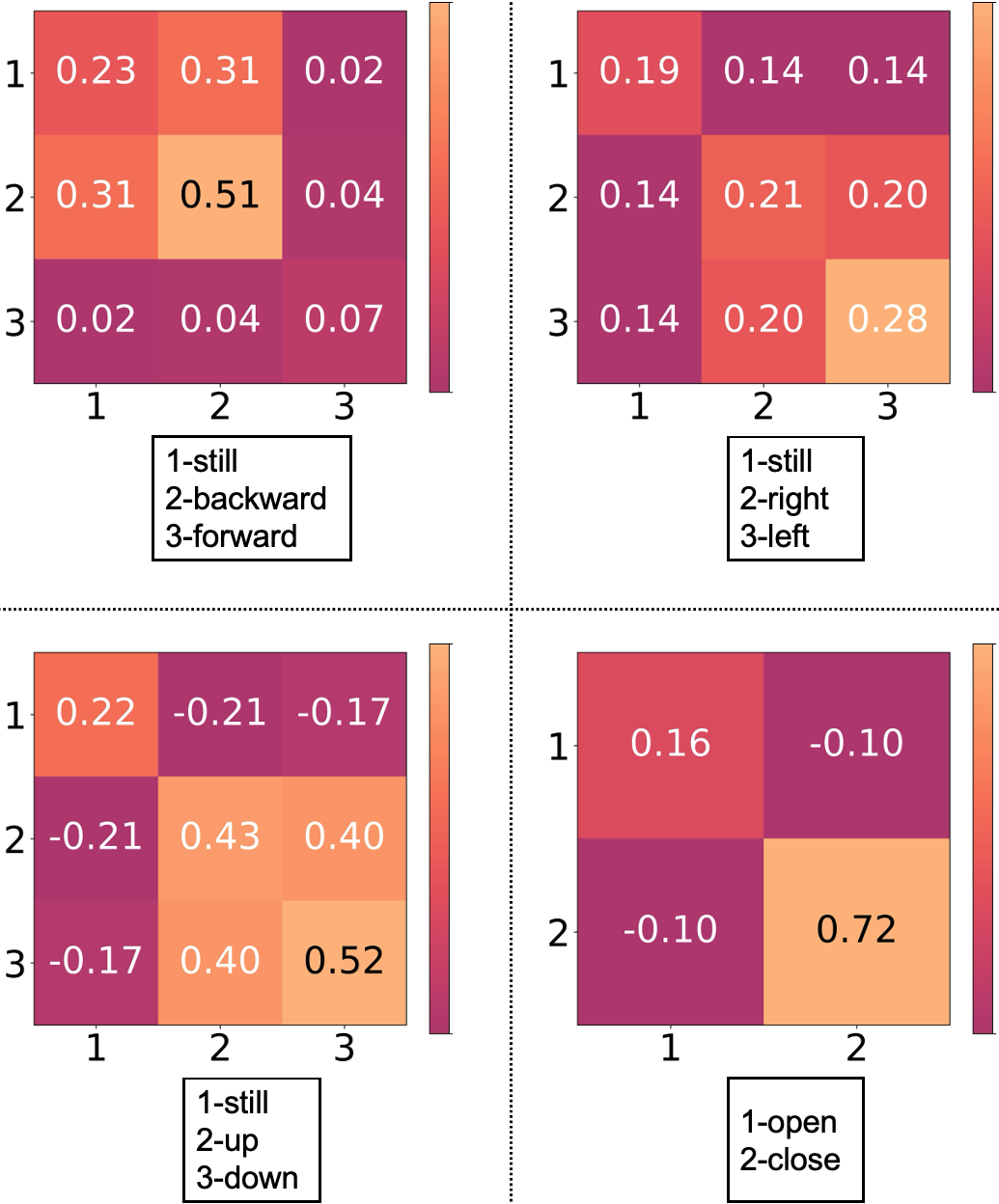}
        \label{fig:motion cluster}
    }
    \hfill
    \subfigure[\textbf{t-SNE Clustering of Manipulation Concept Latents correpsonding to tasks.} We perform t-SNE clustering on the manipulation concepts at each time step. These concepts are generated by our method (Sec.~\ref{sec:method}). Each sample is colored according to its task, representing one of 90 possible tasks as indicated by the colorbar.]{
        \includegraphics[width=0.57\textwidth]{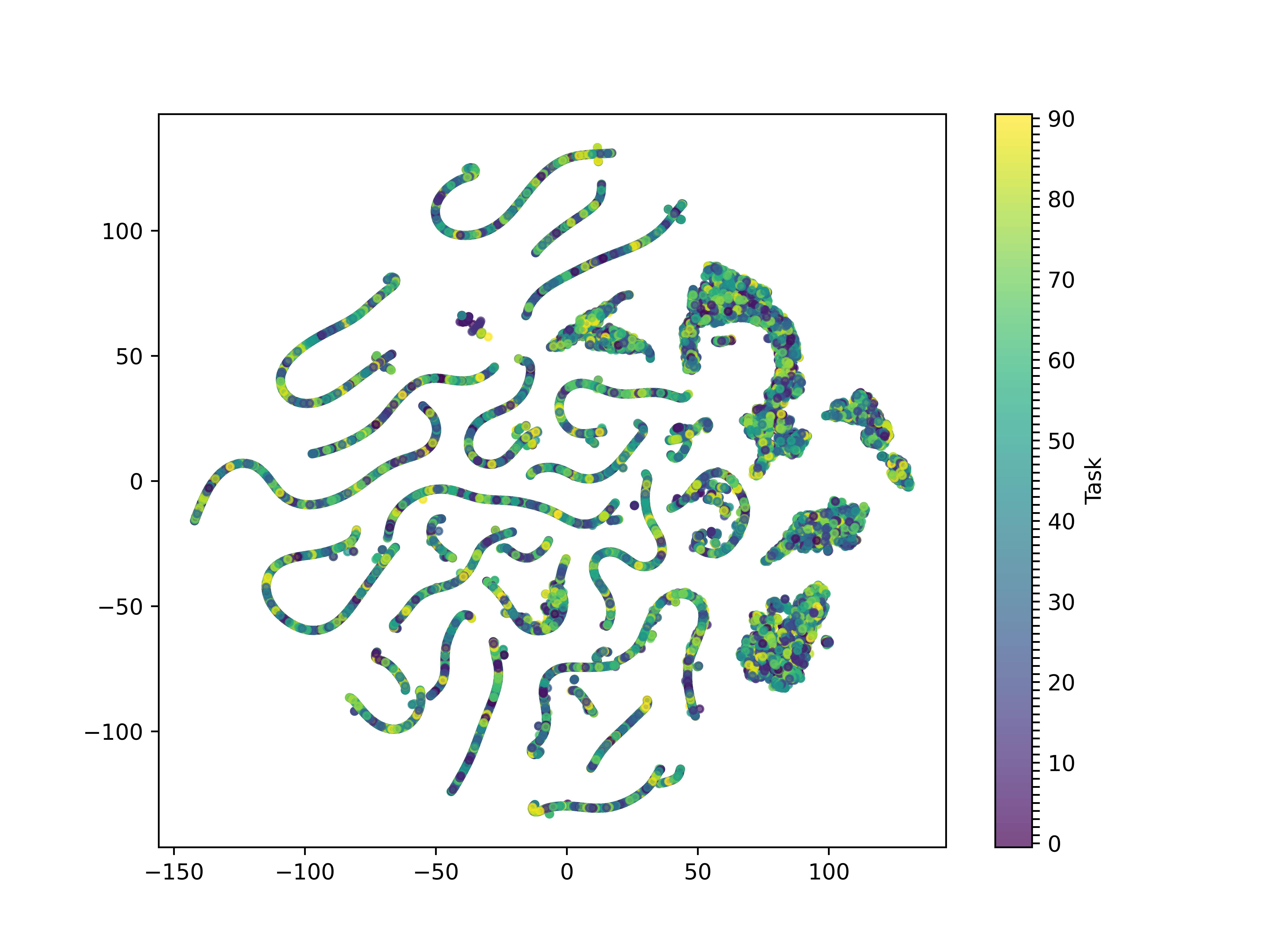}
        \label{fig:tsne-task}
    }
    \caption{}
\end{figure}

We evaluate whether the learned manipulation concept latents capture the robot's motion.
Using Eq.~\ref{eq:class_sim}, we calculate the average similarity ($\times100.0$) between movements based on manipulation concept latents corresponding to specific gripper actions.
Specifically, we collect latents for the following movements from task demonstrations in LIBERO-90:
\begin{itemize}[left=1em,itemsep=0pt,parsep=0pt]
\item [1.] Forward-backward motion: Latents for time-steps where the robot moves forward, backward, or remains still along the forward-backward axis.
\item [2.] Left-right motion: Latents for time-steps where the robot moves left, right, or remains still along the left-right axis.
\item [3.] Up-down motion: Latents for time-steps where the robot moves up, down, or remains still along the up-down axis.
\item [4.] Gripper state: Latents for time-steps where the gripper opens or closes.
\end{itemize}
Movements with velocities below 20\% of the maximum observed velocity are classified as ``still''. Using these collected latents, we generate heatmaps (similar to Fig.~\ref{fig:semantic cluster}) to visualize the average cosine similarity across different movement directions and gripper states (Fig.~\ref{fig:motion cluster}).

The heatmaps reveal that the highest cosine similarity values often appear along the diagonal.
This demonstrates that latents corresponding to the same motion patterns exhibit greater similarity to each other than to those from different motion patterns,
indicating that the latents effectively capture different movement directions and gripper states.
However, we observe that forward-backward motion is captured with lower accuracy compared to other dimensions.
We hypothesize that incorporating additional 3D-informative modalities,
such as depth maps,
beyond the current proprioceptive states could improve the representation of motion along the forward-backward axis.
We leave the exploration of such modality incorporation to future work.

\subsection{Diversity \& Discrimination Study}\label{subsec:diversity and discrimination}
\begin{figure*}[h!]
\centering
\includegraphics[width=0.8\linewidth]{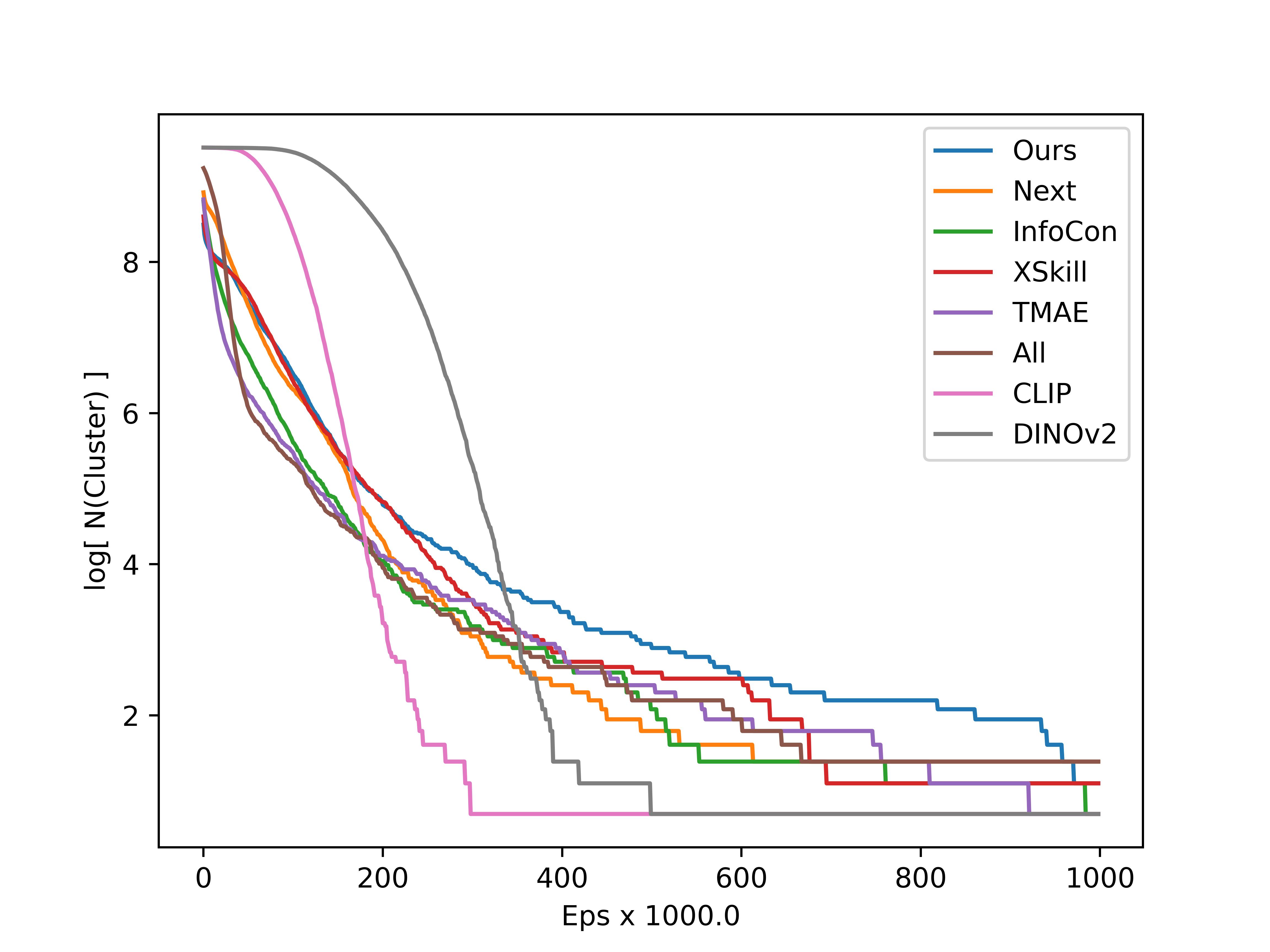}
\caption{
\textbf{DBSCAN Clustering Analysis of Manipulation Concept Latents' Diversity and Discrimination.}
Clustering is performed on manipulation concept latents generated by our method and the baseline methods described in \textbf{Manipulation Concept Discovery Baselines} (Sec.~\ref{subsec:exp_setups}), across 90 tasks from the LIBERO-90 dataset.
The figure shows the (\(\log\)) number of clusters obtained using DBSCAN for clustering density \(\epsilon \in [0, 1]\), with no points classified as noise.
}
\label{fig:dbscan study}
\end{figure*}

We analyze the diversity and discriminability of learned manipulation concepts by comparing concept latents from
our method (Sec.~\ref{sec:method})
and
the baselines in \textit{\textbf{Manipulation Concept Discovery Baselines}} (Sec.~\ref{subsec:exp_setups}).
Specifically,
we cluster latents from these methods and
examine the number of clusters under varying granularities.
The number of clusters reflects concept diversity:
more clusters indicate a wider variety of concepts.
Clustering granularity determines whether clusters are
fine-grained (fine granularity)
or
general (coarse granularity).
Additionally, small granularity perturbations test discriminability,
as less discriminative latents lead to significant clustering changes under small granularity variations.
For each method,
We collect manipulation concept latents from 90 LIBERO-90 tasks (one demonstration per task) and use DBSCAN to cluster them while varying the density parameter \texttt{Eps}, which controls clustering granularity.
Fig.~\ref{fig:dbscan study} shows the cluster counts across different \texttt{Eps} values.
From Fig.~\ref{fig:dbscan study}, our manipulation concept discovery method (\textbf{Ours}) demonstrates two key advantages:
1) At higher granularities (\(\texttt{Eps} > 0.2\)), \textbf{Ours} maintains a \textbf{higher number of clusters}.
2) The \textbf{decline in cluster count is relatively smooth and gradual}, showing stability under small \texttt{Eps} changes.
These results highlight the superior diversity and discriminability of our manipulation concept discovery method.

\subsection{Multi-Level Hierarchical Structure}\label{subsec:multi horizon subgoals}
In Fig.~\ref{fig:vis_hierarchy_example}, we present a visualization example of the \textbf{Multi-Level Hierarchical Structure} described in Sec.~\ref{subsec:direct_concept}. Additional visualization results are available in the supplementary materials under the directory \texttt{supplementary/vis\_multi\_h}.


\subsection{Real Robot Experiments Details}\label{subsec:real-robot-exp-details}

\textit{\textbf{Training Data}}.
As shown in Fig.~\ref{fig:real_exp_setup_main},
the training data for the ``cleaning cup'' task consists of demonstrations using mobile ALOHA \citep{fu2024mobile} to place the cup from the table into the container.
Each demonstration features a scene containing exactly one cup and one container.
There are two pairings of color combinations: 
blue cups with green containers and yellow cups with pink containers.
For each pairing, we collect 27 demonstrations with varied spatial arrangements.

\textit{\textbf{Evaluation Setting}}.
For evaluation, we test our model on six scenarios that introduce variations absent from the training data:

\begin{itemize}[left=0em, itemsep=0pt, parsep=0pt]
    \item \textbf{Novel Placements.} Objects maintain the same color pairings as in training but appear in previously unseen spatial arrangements.
    \item \textbf{Color Composition.} We rearrange color pairings (blue cups with pink containers and yellow cups with green containers) to test generalization across color combinations.
    \item \textbf{Novel Objects.} We introduce unseen objects, such as bamboo-woven containers, pink cups not present in training, or cups initially placed on plates rather than directly on the table.
    \item \textbf{Obstacles.} We position obstacles in front of cups to challenge visual perception.
    \item \textbf{Barriers.} We place a plate inside the container, requiring the robot to lift the cup high enough to clear this barrier when depositing it.
    \item \textbf{Grasping Together.} We position two cups adjacent to one another, requiring the robot to grasp both simultaneously at their contact point and deposit them together in the container.
\end{itemize}

\textit{\textbf{Manipulation Concept Discovery}}.
The model architecture and hyperparameter configuration for manipulation concept discovery follow the methodology described in Sec.~\ref{subsec:id_MCD}.
Since the dataset is relatively small, we adapt smaller transformers: a 4-layer concept encoder (\(\mathcal{E}\), Eq.~\ref{eq:concept_encoder}),
a 4-layer Cross-Modal Correlation Network (\(\mathcal{C}\), Eq.~\ref{eq:loss_mi_and_moc}), and a 4-layer Multi-Horizon Future Predictor (\(\mathcal{F}\), Eq.~\ref{eq:loss_predictable goal}).
For data collected using mobile ALOHA \citep{fu2024mobile},
we incorporate the following modalities: three \(640\times480\) resolution cameras (left-gripper, right-gripper, and upper-gripper)
and 42-dimensional proprioception states (comprising 14-dimensional joint torque, position, and velocity measurements).
All image data undergoes preprocessing as detailed in Sec.~\ref{subsec:id_MCD}.

\textit{\textbf{Enhancing Imitation Learning}}.
Please refer to \textbf{ACT} section in Sec.~\ref{subsec:id_EIL}.

\subsection{Multi-Horizon Goal Prediction Visualization}\label{subsec:vis-pred-and-recon-details}
\begin{figure}[!t]
\centering
\includegraphics[width=1.0\linewidth]{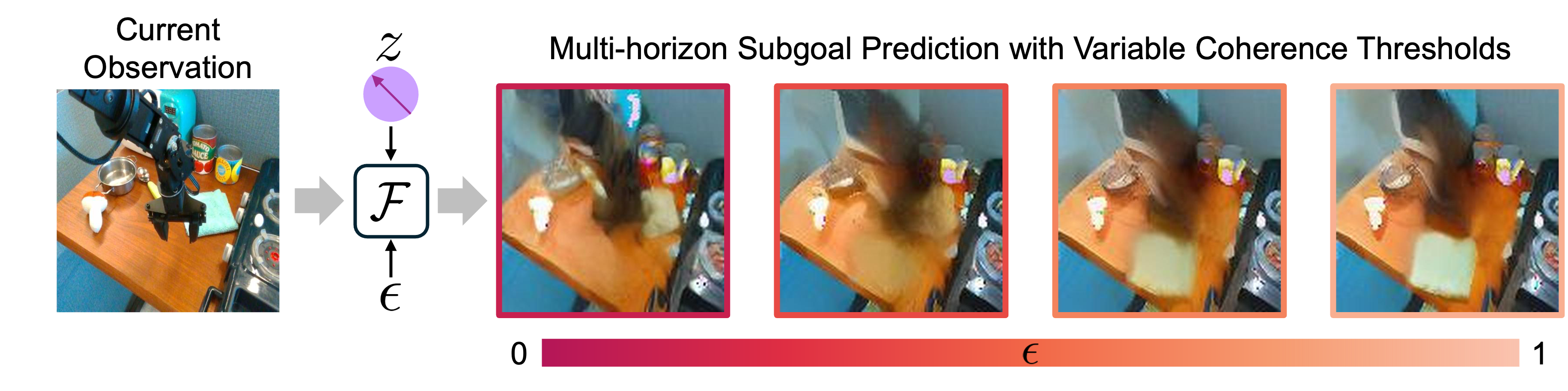}
\caption{
\textbf{Multi-horizon goal prediction with learned manipulation concepts.}
Visualization of future states predicted by our Multi-Horizon Goal Predictor (MHGP, Eq.~\ref{eq:loss_predictable goal}) when conditioned on the current observation, a manipulation concept latent ($z$), and varying coherence thresholds ($\epsilon$). From left to right, as $\epsilon$ increases from 0 to 1, predictions extend progressively further into the future, demonstrating how our manipulation concepts encode temporal abstraction at multiple horizons. Note that predictions capture essential functional relationships (robot-object interactions) rather than pixel-perfect reconstructions, facilitating generalization across environments.
}
\label{fig:pred_and_recon_example}
\end{figure}

We provide visualization results of the \textbf{Multi-Horizon Goal Prediction Visualization}
(Sec.~\ref{subsec:real_world})
in Fig.~\ref{fig:pred_and_recon_example} and supplementary materials under the directory \texttt{supplementary/prediction}.
Below are the details of the experiments:

\paragraph{\textbf{Dataset.}}
For our experiments, we utilized the BridgeDataV2 dataset \citep{walke2023bridgedata}. Since multi-view data is not universally available across all demonstrations, we selected two specific modalities: the robot's proprioceptive states (7DoF) and the third-person camera view. The camera images were preprocessed to \(128\times128\) resolution following the procedure outlined in Sec.~\ref{subsec:id_EIL}.

\paragraph{\textbf{Manipulation Concept Discovery.}}
We implemented the model architecture and hyperparameter configuration as detailed in Sec.~\ref{subsec:id_MCD}, adapting it specifically to operate with the two modalities described in the \textbf{Dataset} section above.

\subsection{Preliminary VLA Integration}

We present a preliminary exploration of integrating manipulation concepts with vision-language-action models (VLAs).
We build upon OpenVLA-OFT \citep{kim2025fine},
which fine-tunes OpenVLA using pretrained parameters and a novel action adapter for downstream tasks.
The action adapter processes hidden layer features from the original pretrained VLA model.
Following this architecture, we introduce an additional ``concept adapter'' that implements the method described in Sec.~\ref{subsec:EIL},
enabling the integration of manipulation concepts into the VLA.

\begin{figure}[ht]
\vspace{-3mm}
\centering
\includegraphics[width=1.0\linewidth]{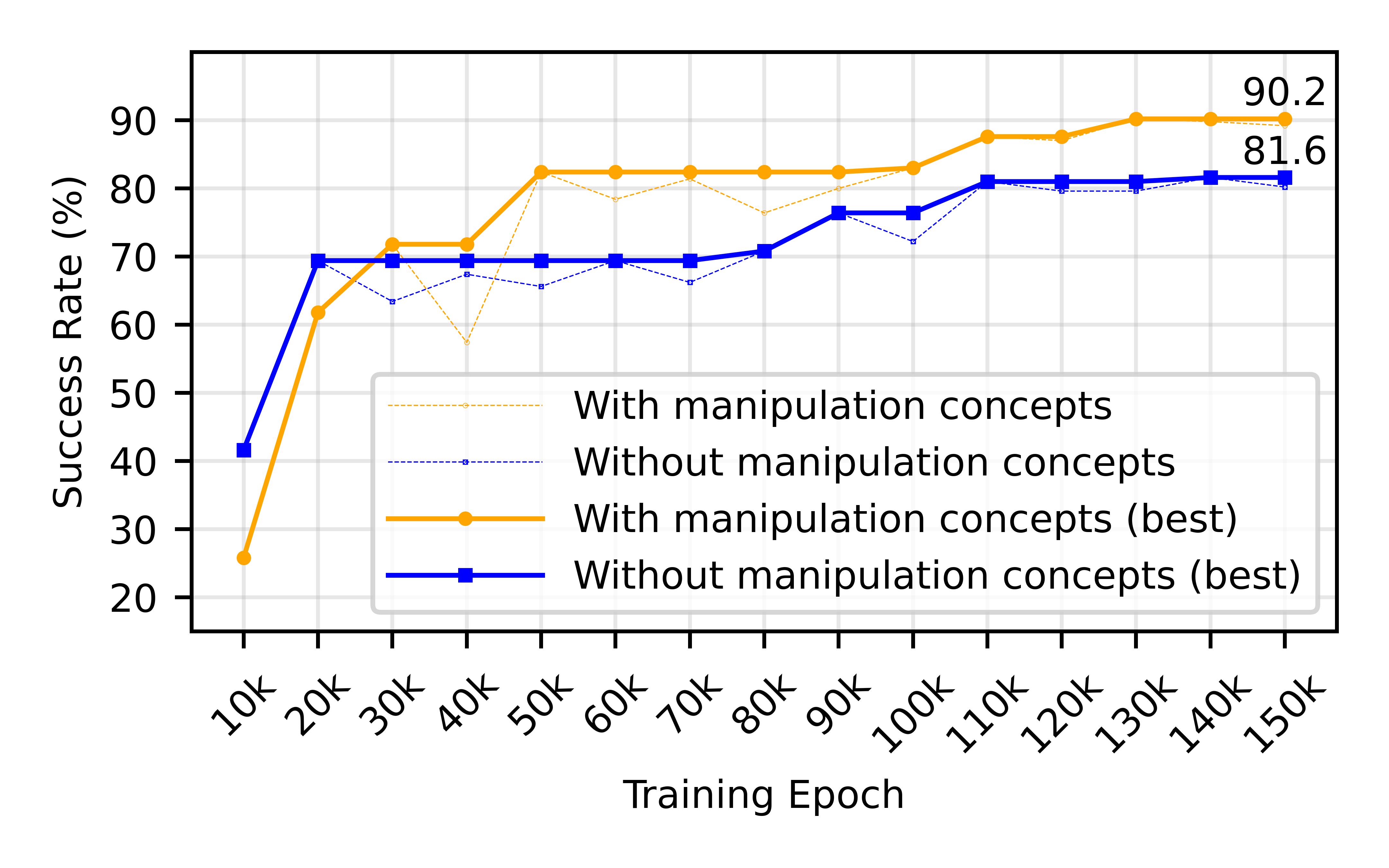}
\vspace{-8mm}
\caption{
Data efficiency comparison on LIBERO-10 tasks with \textbf{50\% training data}.
Solid lines show best performance up to each epoch for models with and without manipulation concepts.
}
\label{fig:vla-data-efficiency}
\end{figure}

To evaluate the data efficiency gains from manipulation concepts,
we fine-tune the enhanced VLA on \textbf{50\% of the training data} used for LIBERO-10 tasks in the original OpenVLA-OFT study \citep{kim2025fine}.
We compare fine-tuning performance with and without manipulation concept integration. Fig.~\ref{fig:vla-data-efficiency} presents the results,
where the x-axis indicates training epochs and the y-axis shows success rates for checkpoints at each epoch.
The solid lines labeled ``best'' represent the highest success rate achieved up to that epoch.

The results demonstrate that manipulation concepts improve data utilization.
With only \textbf{half the training data}, the concept-enhanced approach consistently achieves higher success rates throughout training. Notably, the original OpenVLA-OFT achieved 94.5\% success with the full dataset \citep{kim2025fine}, while our concept-enhanced model with \textbf{half the data} reaches comparable performance levels, indicating substantially improved data efficiency.

We hypothesize that this improvement stems from HiMaCon's ability to capture manipulation dynamics at multiple abstraction levels.
The learned concepts provide explicit intermediate representations that bridge high-level task instructions and low-level control actions,
thereby reducing the learning burden on VLAs by supplying structured manipulation knowledge rather than requiring learning of complex sensorimotor patterns from scratch.
Further investigation of this integration will be pursued in future work.

\section{Limitations \& Future works}\label{subsec:future works}
\textbf{Further Exploration of multi-modality.}
We propose enhancing robotic data collection with richer modalities and studying how these modalities can derive more effective manipulation concepts. While current robotics research primarily focuses on visual information, human manipulation relies on multiple sensory inputs, particularly tactile feedback to complement vision. This is especially crucial for robotic systems with limited tactile capabilities. Future work should investigate which modalities contribute most significantly to performance improvements and how to fully leverage their potential.

\textbf{Further Exploration of multi-horizon sub-goal.}
Our work proposes methods to derive sub-processes for achieving sub-goals across multiple horizons, though several improvements remain possible. Current methods inadequately capture relationships between different values of \(\epsilon\) in Eq.~\ref{eq:t_structure}, failing to reflect the natural tree structure of hierarchical sub-goals. Future research could explicitly derive tree structures \citep{wan2024lotus,zhu2022bottom} where long-horizon sub-goals serve as parent nodes to short-horizon child nodes. Additionally, our cosine similarity approach for determining sub-goal correspondence could be refined with more sophisticated metrics.


\textbf{Scaling up.}
Computational constraints have limited our exploration of how our method scales with larger datasets. We plan to leverage pretrained multi-modal foundation models, adopting structures inspired by \citep{bruce2024genie} and extending pretraining beyond robotics data as in \citep{ye2024latent}. We also aim to further investigate whether our manipulation concepts can enhance advanced policies like Vision-Language-Action models \citep{black2024pi_0,intelligence2025pi,kim2025fine,kim2024openvla}.




\begin{algorithm}[h!]
\caption{Derive Subprocess \(\mathrm{h}(\mathbf{z}_i; \epsilon)\)}
\label{alg:e_cluster}
\begin{algorithmic}
\STATE {\bfseries Input:} manipulation concept vectors $\mathbf{z}_i=\{z_i^t\}_{t=1}^{T_i}$,
coherence parameter $\epsilon\in[0,1]$.
\STATE {\bfseries Initialize:} $End=[]$, $g_b=1$
\WHILE {$g_b \leq T_i$}
    \STATE $g_e = g_b + 1$
    \WHILE {$true$}
        \IF{$\exists u\in[g_b,g_e),\text{s.t. }\text{dist}(z_i^u,z_i^{g_e})\geq\epsilon$
        \textbf{or}
        $g_e > T_i$}
            \STATE \textbf{break}
        \ENDIF
        \STATE $g_e=g_e+1$
    \ENDWHILE
    \STATE  $End.append\left([g_b,g_e)\right)$
    \STATE  $g_b = g_e$
\ENDWHILE
\STATE \textbf{Return} $End$
\end{algorithmic}
\end{algorithm}

\begin{algorithm}[h!]
\caption{Manipulation Concept Discovery Training
(one demonstration per batch)}
\label{alg:mcd_pseudocode}
\begin{algorithmic}
\STATE {\bfseries Input:} demonstrations $\mathbf\tau_i\in D$,
where
$\tau_i=
\{(o^{1,t}_i,o^{2,t}_i,...,o^{M,t}_i,a^t_i)\}_{t=1}^{T_i}$
\STATE {\bfseries Initialize:}
Manipulation concept assignment encoder $\mathcal{E}(\cdot;\Theta_\mathcal{E})$
\STATE {\bfseries Initialize:}
Modality Correlation Learner $\mathcal{C}(\cdot;\Theta_c)$,
Subgoal Learner $\mathcal{F}(\cdot;\Theta_f)$
\WHILE{$true$}
\FOR{$\tau_i$ {\bfseries in} $D$}
    \STATE $(z_i^1,\cdots,z_i^{T_i})\leftarrow\mathcal{E}\left((o^{1,1}_i,...,o^{M,1}_i),(o^{1,2}_i,...,o^{M,2}_i),\cdots,(o^{1,T_i}_i,...,o^{M,T_i}_i);\Theta_\mathcal{E}\right)$
    \WHILE{True}
        \STATE Randomly generate a tuple $(m_1, m_2, \ldots, m_M)$, where $m_i \in \{0, 1\}$
        \IF{$\sum_{i=1}^M m_i < M$}
            \STATE \textbf{break}
        \ENDIF
    \ENDWHILE
    \STATE $(\hat{o}_i^{1,t},\cdots,\hat{o}_i^{M,t})_{t=1}^{T_i}\leftarrow\mathcal{C}\left((o^{1,t}_i\cdot m_1,o^{2,t}_i\cdot m_2,...,o^{M,t}_i\cdot m_M, z_i^t)_{t=1}^{T_i};\Theta_c\right)$
    \STATE
    $\mathcal{L}_\text{mm}=\sum_{t=1}^{T_i}\sum_{m=1}^M
    \left\| \hat{o}_i^{m,t}-o_i^{m,t} \right\|$
    \STATE $\epsilon\sim\mathrm{U}([0,1])$
    \STATE $End=\mathrm{h}(z_i^1,\cdots,z_i^{T_i};\epsilon)$
    \COMMENT{Alg.~\ref{alg:e_cluster}}
    \FOR{$t=1$ {\bfseries to} $T_i$}
        \STATE $\mathbf{g}_t=\min\left(\big\{g_e\mid[g_b,g_e)\in End,g_e>t\}\cup\{T_i\big\}\right)$
    \ENDFOR
    \STATE $(\overline{o}_i^{1,t},\cdots,\overline{o}_i^{M,t})_{t=1}^{T_i}\leftarrow\mathcal{F}\left((o^{1,t}_i,o^{2,t}_i,...,o^{M,t}_i,z_i^t,\epsilon)_{t=1}^{T_i};\Theta_f\right)$
    \STATE
    $\mathcal{L}_\text{mh}=\sum_{t=1}^{T_i}\sum_{m=1}^M
    \left\| \overline{o}_i^{m,t}-o_i^{m,\mathbf{g}_t} \right\|$
\ENDFOR
\ENDWHILE
\end{algorithmic}
\end{algorithm}

\clearpage



\clearpage

\clearpage

\clearpage
\section*{NeurIPS Paper Checklist}

The checklist is designed to encourage best practices for responsible machine learning research, addressing issues of reproducibility, transparency, research ethics, and societal impact. Do not remove the checklist: {\bf The papers not including the checklist will be desk rejected.} The checklist should follow the references and follow the (optional) supplemental material.  The checklist does NOT count towards the page
limit. 

Please read the checklist guidelines carefully for information on how to answer these questions. For each question in the checklist:
\begin{itemize}
    \item You should answer \answerYes{}, \answerNo{}, or \answerNA{}.
    \item \answerNA{} means either that the question is Not Applicable for that particular paper or the relevant information is Not Available.
    \item Please provide a short (1–2 sentence) justification right after your answer (even for NA). 
\end{itemize}

{\bf The checklist answers are an integral part of your paper submission.} They are visible to the reviewers, area chairs, senior area chairs, and ethics reviewers. You will be asked to also include it (after eventual revisions) with the final version of your paper, and its final version will be published with the paper.

The reviewers of your paper will be asked to use the checklist as one of the factors in their evaluation. While "\answerYes{}" is generally preferable to "\answerNo{}", it is perfectly acceptable to answer "\answerNo{}" provided a proper justification is given (e.g., "error bars are not reported because it would be too computationally expensive" or "we were unable to find the license for the dataset we used"). In general, answering "\answerNo{}" or "\answerNA{}" is not grounds for rejection. While the questions are phrased in a binary way, we acknowledge that the true answer is often more nuanced, so please just use your best judgment and write a justification to elaborate. All supporting evidence can appear either in the main paper or the supplemental material, provided in appendix. If you answer \answerYes{} to a question, in the justification please point to the section(s) where related material for the question can be found.

IMPORTANT, please:
\begin{itemize}
    \item {\bf Delete this instruction block, but keep the section heading ``NeurIPS Paper Checklist"},
    \item  {\bf Keep the checklist subsection headings, questions/answers and guidelines below.}
    \item {\bf Do not modify the questions and only use the provided macros for your answers}.
\end{itemize}


\begin{enumerate}

\item {\bf Claims}
    \item[] Question: Do the main claims made in the abstract and introduction accurately reflect the paper's contributions and scope?
    \item[] Answer: \answerYes{} 
    \item[] Justification: The abstract gives a summary of our contribution on self-supervised learning of manipulation concepts.

\item {\bf Limitations}
    \item[] Question: Does the paper discuss the limitations of the work performed by the authors?
    \item[] Answer: \answerYes{} 
    \item[] Justification: We discuss limitations including improvements to hierarchy derivation, further work on scaling up, and modality balance.

\item {\bf Theory assumptions and proofs}
    \item[] Question: For each theoretical result, does the paper provide the full set of assumptions and a complete (and correct) proof?
    \item[] Answer: \answerNA{} 
    \item[] Justification: We mainly make use of established theoretical frameworks (such as mutual information) for clarification and modeling of our method.

    \item {\bf Experimental result reproducibility}
    \item[] Question: Does the paper fully disclose all the information needed to reproduce the main experimental results of the paper to the extent that it affects the main claims and/or conclusions of the paper (regardless of whether the code and data are provided or not)?
    \item[] Answer: \answerYes{} 
    \item[] Justification: We provide details in the appendix and supplementary materials.

\item {\bf Open access to data and code}
    \item[] Question: Does the paper provide open access to the data and code, with sufficient instructions to faithfully reproduce the main experimental results, as described in supplemental material?
    \item[] Answer: \answerNo{} 
    \item[] Justification: We will open source all code and newly-created datasets upon acceptance.

\item {\bf Experimental setting/details}
    \item[] Question: Does the paper specify all the training and test details (e.g., data splits, hyperparameters, how they were chosen, type of optimizer, etc.) necessary to understand the results?
    \item[] Answer: \answerYes{} 
    \item[] Justification: Details are specified in the Experiments section and in the appendix and supplementary materials.

\item {\bf Experiment statistical significance}
    \item[] Question: Does the paper report error bars suitably and correctly defined or other appropriate information about the statistical significance of the experiments?
    \item[] Answer: \answerNo{} 
    \item[] Justification: For the policy success rates we currently include standard deviation.

\item {\bf Experiments compute resources}
    \item[] Question: For each experiment, does the paper provide sufficient information on the computer resources (type of compute workers, memory, time of execution) needed to reproduce the experiments?
    \item[] Answer: \answerYes{} 
    \item[] Justification: Please refer to the details provided in the appendix.
    
\item {\bf Code of ethics}
    \item[] Question: Does the research conducted in the paper conform, in every respect, with the NeurIPS Code of Ethics \url{https://neurips.cc/public/EthicsGuidelines}?
    \item[] Answer: \answerYes{} 
    \item[] Justification: Current experiments and topics do not conflict with the Code of Ethics.

\item {\bf Broader impacts}
    \item[] Question: Does the paper discuss both potential positive societal impacts and negative societal impacts of the work performed?
    \item[] Answer: \answerNA{} 
    \item[] Justification: Currently, the experiments are carried out in simulations and on robots in laboratories.
    
\item {\bf Safeguards}
    \item[] Question: Does the paper describe safeguards that have been put in place for responsible release of data or models that have a high risk for misuse (e.g., pretrained language models, image generators, or scraped datasets)?
    \item[] Answer: \answerNA{} 
    \item[] Justification: Currently, we have not encountered any safeguard issues.

\item {\bf Licenses for existing assets}
    \item[] Question: Are the creators or original owners of assets (e.g., code, data, models), used in the paper, properly credited and are the license and terms of use explicitly mentioned and properly respected?
    \item[] Answer: \answerYes{} 
    \item[] Justification: We have checked the sources we used.

\item {\bf New assets}
    \item[] Question: Are new assets introduced in the paper well documented and is the documentation provided alongside the assets?
    \item[] Answer: \answerNA{} 
    \item[] Justification: We will release all new assets we created (code/models/datasets) upon acceptance.

\item {\bf Crowdsourcing and research with human subjects}
    \item[] Question: For crowdsourcing experiments and research with human subjects, does the paper include the full text of instructions given to participants and screenshots, if applicable, as well as details about compensation (if any)? 
    \item[] Answer: \answerNA{} 
    \item[] Justification: We currently do not have crowdsourcing experiments.

\item {\bf Institutional review board (IRB) approvals or equivalent for research with human subjects}
    \item[] Question: Does the paper describe potential risks incurred by study participants, whether such risks were disclosed to the subjects, and whether Institutional Review Board (IRB) approvals (or an equivalent approval/review based on the requirements of your country or institution) were obtained?
    \item[] Answer:  \answerNA{} 
    \item[] Justification: We currently do not involve crowdsourcing nor research with human subjects.

\item {\bf Declaration of LLM usage}
    \item[] Question: Does the paper describe the usage of LLMs if it is an important, original, or non-standard component of the core methods in this research? Note that if the LLM is used only for writing, editing, or formatting purposes and does not impact the core methodology, scientific rigorousness, or originality of the research, declaration is not required.
    \item[] Answer: \answerNA{}. 
    \item[] Justification: The core method development in this research does not involve LLMs.

\end{enumerate}




\end{document}